%% file: neurips_2020.tex
\documentclass{article}

\usepackage[final,nonatbib]{neurips_2020}

\usepackage[utf8]{inputenc} % allow utf-8 input
\usepackage[T1]{fontenc}    % use 8-bit T1 fonts
\usepackage{hyperref}       % hyperlinks
\usepackage{url}            % simple URL typesetting
\usepackage{booktabs}       % professional-quality tables
\usepackage{amsfonts}       % blackboard math symbols
\usepackage{nicefrac}       % compact symbols for 1/2, etc.
\usepackage{microtype}      % microtypography
\usepackage[numbers]{natbib}

\input{packages}

\title{A Causal View on Robustness  of Neural Networks}

\author{%
   Cheng Zhang \thanks{Equal contribution} \\\hspace{-10pt}
   Microsoft Research \\\hspace{-10pt}
   \texttt{Cheng.Zhang@microsoft.com} \\ \hspace{-10pt}
     \And
   Kun Zhang \\ \hspace{-10pt}
   Carnegie Mellon University \\ \hspace{-10pt}
   \texttt{kunz1@cmu.edu} \\\hspace{-10pt}
     \And
   Yingzhen Li \footnotemark[1] \\
   Microsoft Research \\
   \texttt{Yingzhen.Li@microsoft.com} \\
}

\begin{document}

\maketitle
%\vspace{-10pt}
\input{abstract.tex}
\input{intro.tex}

\input{causal_view.tex}

\input{model.tex}

\input{exp.tex}

\input{related.tex}
\input{discussion.tex}
\bibliography{ref}
\bibliographystyle{plainnat}

\clearpage
\input{appendix.tex}

\end{document}

%% file: packages.tex
\usepackage{microtype}
\usepackage{graphicx}
\usepackage{subfigure}
\usepackage{booktabs} % for professional tables

\usepackage{hyperref}

\usepackage{amsmath, amssymb}
\usepackage{graphicx}
\usepackage{subfigure}
\usepackage{multirow}
\usepackage{color,comment,rotating,subfigure,url,xspace} % author packages
\usepackage{tikz}
\usetikzlibrary{arrows,shadows,shapes,backgrounds,decorations,snakes,fit}
\usepackage{algorithm}
\usetikzlibrary{shadows}

\usepackage{paralist}
\usepackage{wrapfig}
\usepackage{lipsum}
\usepackage[font=small,skip=5pt]{caption}

\usepackage{natbib}

%% file: abstract.tex
\begin{abstract}
We present a causal view on the robustness of neural networks against input manipulations, which applies not only to traditional classification tasks but also to general measurement data. Based on this view, we design a deep causal manipulation augmented model (deep CAMA) which explicitly models possible manipulations on certain causes leading to changes in the observed effect.  We further develop data augmentation and test-time fine-tuning methods to improve deep CAMA's robustness. When compared with discriminative deep neural networks, our proposed model shows superior robustness against unseen manipulations. As a by-product, our model achieves disentangled representation which separates the representation of manipulations from those of other latent causes.
\end{abstract}

%% file: intro.tex
\vspace{-10pt}
\section{Introduction}
\label{sec:intro}

\begin{wrapfigure}{r}{0.55\textwidth}
    %\centering
    \vspace{-36pt}
    \subfigure[Test Vertical shift]{
    \includegraphics[width = 0.47\linewidth, bb= 0 0 430 430]{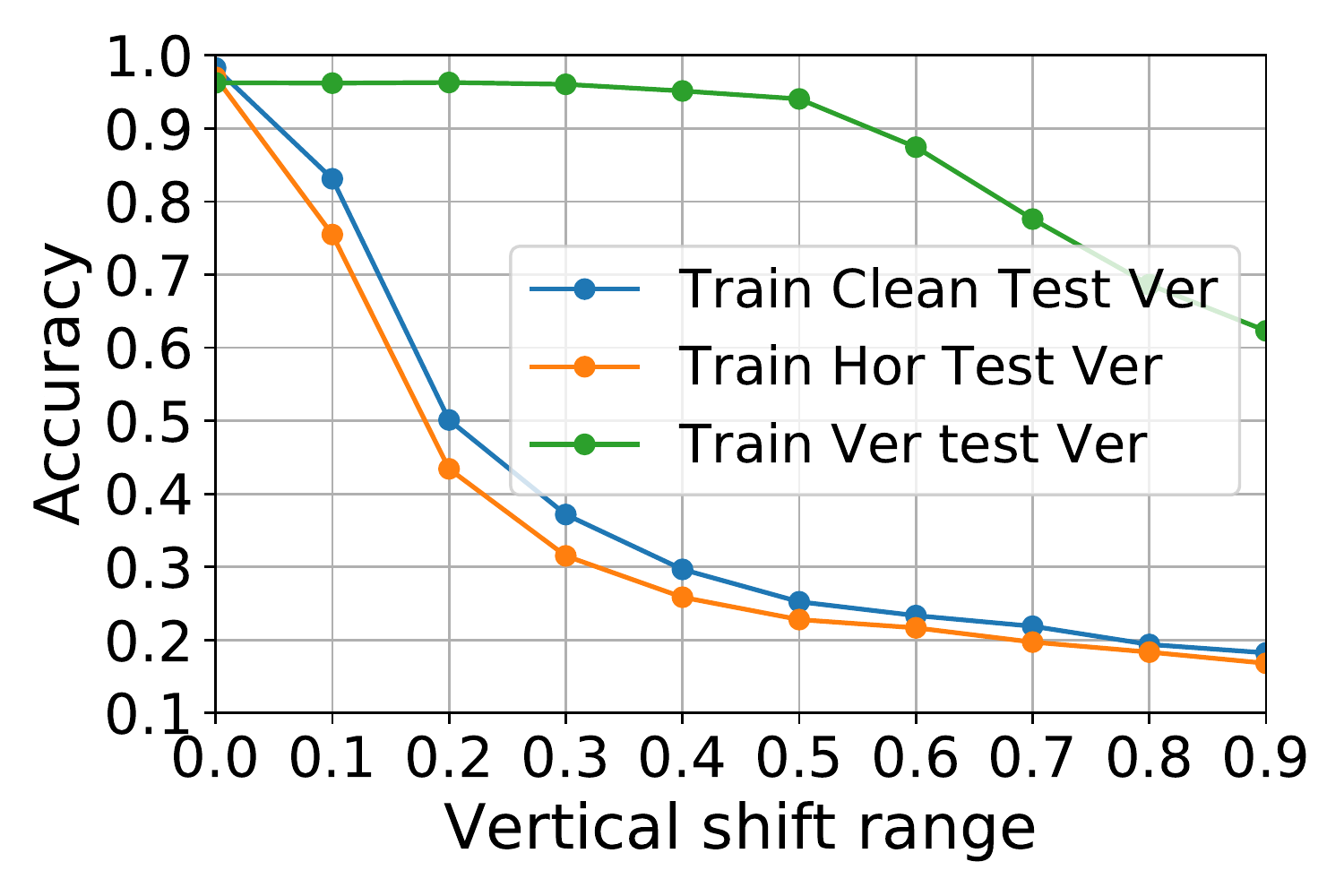}
    \label{fig:disc_vt}
    }
    \hspace{-5pt}
    \subfigure[Test Horizontal shift]{
    \includegraphics[width = 0.47\linewidth, bb= 0 0 430 430]{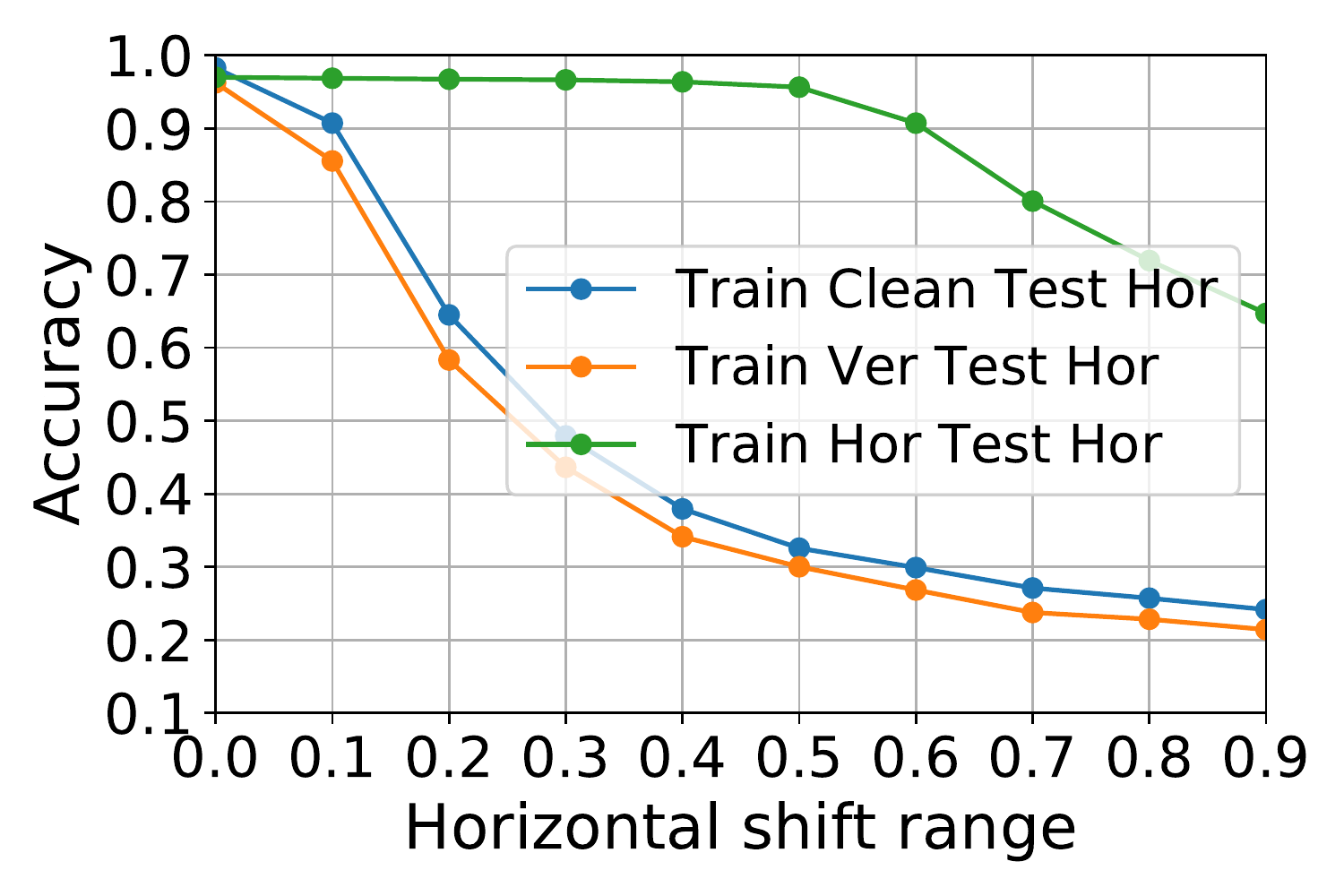}
    \label{fig:disc_ht}
    }
    \caption{
    Robustness of DNNs on shifted MNIST. Panels (a) and (b) show the accuracy on classifying noisy test data generated by shifting the digits vertically (Ver) and horizontally (Hor). It shows that data augmentation during training makes generalization to unseen shifts worse (orange vs blue lines).
    }
        \label{fig:disc_test}
    \vspace{-12pt}
%\end{figure}
\end{wrapfigure}

Deep neural networks (DNNs) have great success in many real-life applications; however, they are easily fooled even by a tiny amount of perturbation \citep{szegedy2013intriguing,goodfellow:explaining2015,carlini:bypass2017,athalye:obfuscated2018}. Lack of robustness hinders the application of DNNs to critical decision making tasks such as uses in healthcare. 
To address this, one may suggest training DNNs with diverse datasets. Indeed, data augmentation and adversarial training have shown improvements in both the generalization and robustness of DNNs \citep{kurakin2016adversarial,perez2017effectiveness,madry2017towards}. Unfortunately, this does not address the vulnerability of DNNs to \emph{unseen} manipulations, e.g.~as shown in Figure \ref{fig:disc_test}, a DNN trained on clean MNIST digits fails to properly classify shifted digits. Although observing perturbations of clean data in training improves robustness against that particular manipulation (the green line), the DNN is still fragile when unseen manipulations are present (orange line).
As it is unrealistic to augment the training data towards all possible manipulations that might occur, a principled method that fundamentally improves the robustness is much needed.

%\begin{figure}[t]

On the other hand, human perception is robust to such perturbations thanks to the capability of \emph{causal reasoning} \citep{pearl2018book,gopnik2004theory}. 
After learning the concept of an ``elephant'', a child can identify the elephant in a photo taken under any lighting condition, location, etc. In the causal view, the lighting condition and the location are causes of the presented scene, which can be intervened without changing the presence of the elephant. However, discriminative DNNs do not take such possible interventions into account, and cannot adapt the predictor for new data gathered with unseen manipulation. 

In light of the above observation, we discuss the robustness of DNNs from a causal perspective. Our contributions are:
\begin{compactitem}
\item \textbf{A causal view on robustness of neural networks.} %
%We argue from a causal perspective that adversarial examples for a model can be generated by manipulations on the effect variables and/or their unseen causes. Therefore DNN's vulnerability to adversarial attacks is due to the lack of causal reasoning.
We argue from a causal perspective that input perturbations are generated by their unseen causes that are \emph{artificially manipulatable}. Therefore DNN's robustness issues to these input perturbations is due to the lack of causal understanding.\vspace{0.5em}

\item \textbf{A causal inspired deep generative model.}
We design a causal inspired deep generative model which takes into account possible  interventions on the causes in the data generation process \cite{woodward2005making}.  
Accompanied with this model is a test-time inference method to learn unseen interventions and thus improve classification accuracy on manipulated data. 
Compared to DNNs, experiments on both MNIST and a measurement-based dataset show that our model is significantly more robustness to unseen manipulations. 
\end{compactitem}

%% file: causal_view.tex
\vspace{-3pt}
\section{A Causal View on Robustness of Neural Networks}
\label{sec:causalView}
\vspace{-2pt}
Discriminative DNNs may not be robust to manipulations such as adversarial noise injection \citep{goodfellow:explaining2015,carlini:attack2017,athalye:obfuscated2018}, rotation, and shift. 
They simply trust the observed data and ignore the constraints of the data generating process,
which leads to overfiting to nuisance factors and makes the classification output sensitive to such factors. By exploiting the overfit to the nuisance factors, an adversary can easily manipulate the inputs to fool discriminative DNNs into predicting the wrong outcomes. 

On the contrary, human can easily recognize an object in a scene and be indifferent to the variations in other aspects such as background, viewing angle, the presence of a sticker to the object, etc. More importantly, human recognition is less affected even by drastic perturbations on a number of factors underlying the observed data, e.g.~variations in the lighting condition. We argue that the main difference here is due to our ability to perform \emph{causal reasoning}, which identifies factor that are not relevant to the recognition results \citep{freeman1994generic,peters2017elements,parascandolo2017learning}. This leads to robust human perception to not only a certain type of perturbations, but also to many types of unseen manipulations on other factors. Thus we argue that one should incorporate the causal perspective into model design, and make the model robust on the level of different types of manipulations.

Before presenting our causally informed model, we first define the generative process of perceived data. There might exist multiple causes in the data generation process influencing the observed data $X$, and we visualize exemplar causal graphs in Figure \ref{fig:example_all} with the arrows indicating causal associations. Among these causes of $X$, $Y$ is the target to be predicted, $M$ is a set of variables which can be intervened artificially, and $Z$ represents the rest of the causes that cannot be intervened in the application context. Take hand-written digit classification for example, $X$ is the image and $Y$ is the class label. The appearance of $X$ is an effect of the digit number $Y$, latent causes $Z$ such as writing styles, and possible manipulations $M$ such as rotation or translation.

We can thus define valid perturbations of data through the lens of causality. Datasets are produced by interventions in general, so defining a valid attack is equivalent to defining a set of variables in the causal graph (Figure \ref{fig:example_all}) which can be intervened by the adversary. We argue that \emph{a valid perturbation is an intervention on $M$ which, together with the original $Y$ and $Z$, produces the manipulated data $X$}. In this regard, recent adversarial attacks as perturbations of the inputs can be considered as a specific type of intervention on $M$ such that a learned predictor is deceived.
Note here we do not consider interventions on $Y$ and $Z$ (and their causes): interventions on $Y$ (and its causes) changes the ``true" value of the target and do not correspond to the type of perturbations we are considering; by definition $Z$ (and its causes) cannot be intervened \emph{artificially} (e.g.~genetic causes are often difficult to intervene), thereby unavailable to the adversary. This also shows the importance of separating the (unobserved) causes $Z$ and $M$, as it helps to better identify the interventions presented in perturbed inputs, which also leads to improved classification robustness.

In light of the above definition on valid perturbations, it is clear that performing prediction adaptive to the (unknown) intervention is necessary to achieve robustness to manipulated data. 
A natural way to build such adaptive predictor is to construct a model that perform reasoning in a  way consistent to the causal process. To see this, note that a valid perturbation changes the value of $M$, but it leaves the functional relationship from $M$ and $Y$ to $X$ intact. This is known as \emph{modularity property} \cite{woodward2005making}, and in this sense the causal system is autonomous \cite{pearl2009causality}. Therefore a causally consistent predictive model is expected to be able to learn this functional relationship from data, and adapt the prediction result of target in test time according to its reasoning on the underlying causal factors.

\begin{figure}
\begin{minipage}[]{0.59\textwidth}
\vspace{-20pt}
    \centering
    \subfigure[factorized \label{fig:example}]{ \scalebox{0.6}{\input{tikz_eg.tex}}}
    \subfigure[ $Y$ causes $M$ \label{fig:example_y_cause_m}]{
    \scalebox{0.6}{\input{tikz_YM.tex}}
    }
    \subfigure[confounded \label{fig:example_confounding}]{
    \scalebox{0.6}{\input{tikz_confound.tex}}
    }
    \vspace{-5pt}
     \caption{Exampler causal graphs with $Y$, $Z$, $M$ causing $X$. $Y$ might cause $M$ (panel b), or they might be confounded (panel c). }
    \label{fig:example_all}
    \vspace{-15pt}
\end{minipage}
\hspace{15pt}
\begin{minipage}[]{0.33\textwidth}
\centering
\scalebox{0.8}{\input{tikz_model.tex}}
\vspace{2pt}
\caption{Graphical presentation of deep CAMA for single modal data. }
%\vspace{-5pt}
\label{fig:CAMA1}
\end{minipage}
%\vspace{-10pt}
\end{figure}

%% file: tikz_eg.tex
\pgfdeclarelayer{background}
\pgfdeclarelayer{foreground}
\pgfsetlayers{background,main,foreground}

\begin{tikzpicture}

\tikzstyle{scalarnode} = [circle, draw, fill=white!11,  
    text width=1.2em, text badly centered, inner sep=2.5pt]
\tikzstyle{arrowline} = [draw,color=black, -latex]
;
%nodes
    \node [scalarnode] at (0,0) (X)   {$X$};
    \node [scalarnode] at (-1.5, 1.5) (Y) {$Y$};
    \node [scalarnode] at (0, 1.5) (Z) {$Z$};
    \node [scalarnode] at (1.5, 1.5) (M) {$M$};
%line
    \path [arrowline]  (Y) to (X);
    \path [arrowline]  (Z) to (X);
    \path [arrowline]  (M) to (X);
\end{tikzpicture}

%% file: tikz_YM.tex
\pgfdeclarelayer{background}
\pgfdeclarelayer{foreground}
\pgfsetlayers{background,main,foreground}

\begin{tikzpicture}

\tikzstyle{scalarnode} = [circle, draw, fill=white!11,  
    text width=1.2em, text badly centered, inner sep=2.5pt]
\tikzstyle{arrowline} = [draw,color=black, -latex]
;
%nodes
    \node [scalarnode] at (0,0) (X)   {$X$};
    \node [scalarnode] at (-1.5, 1.5) (Y) {$Y$};
    \node [scalarnode] at (0, 1.5) (Z) {$Z$};
    \node [scalarnode] at (1.5, 1.5) (M) {$M$};
%line
    \path [arrowline]  (Y) to (X);
    \path [arrowline]  (Z) to (X);
    \path [arrowline]  (M) to (X);
    \path [arrowline]  (Y) to[out=75,in=115, distance=0.5cm ] (M);
\end{tikzpicture}

%% file: tikz_confound.tex
\pgfdeclarelayer{background}
\pgfdeclarelayer{foreground}
\pgfsetlayers{background,main,foreground}

\begin{tikzpicture}

\tikzstyle{scalarnode} = [circle, draw, fill=white!11,  
    text width=1.2em, text badly centered, inner sep=2.5pt]
\tikzstyle{arrowline} = [draw,color=black, -latex]
;
%nodes
    \node [scalarnode] at (0,0) (X)   {$X$};
    \node [scalarnode] at (-1.5, 1.5) (Y) {$Y$};
    \node [scalarnode] at (0, 1.5) (Z) {$Z$};
    \node [scalarnode] at (1.5, 1.5) (M) {$M$};
    
    \node [scalarnode] at (0, 3) (L) {$L$};
%line
    \path [arrowline]  (Y) to (X);
    \path [arrowline]  (Z) to (X);
    \path [arrowline]  (M) to (X);
    
    \path [arrowline]  (L) to (M);
    \path [arrowline]  (L) to (Y);
\end{tikzpicture}

%% file: tikz_model.tex
\pgfdeclarelayer{background}
\pgfdeclarelayer{foreground}
\pgfsetlayers{background,main,foreground}

\begin{tikzpicture}

\tikzstyle{scalarnode} = [circle, draw, fill=white!11,  
    text width=1.2em, text badly centered, inner sep=2.5pt]
\tikzstyle{arrowline} = [draw,color=black, -latex]
;
\tikzstyle{dasharrowline} = [draw,dashed, color=black, -latex]
;
\tikzstyle{surround} = [thick,draw=black,rounded corners=1mm]
%nodes
    \node [scalarnode, fill=black!30] at (0,0) (X)   {$X$};
    \node [scalarnode, fill=black!30] at (-1.5, 1.5) (Y) {$Y$};
    \node [scalarnode] at (0, 1.5) (Z) {$Z$};
    \node [scalarnode] at (1.5, 1.5) (M) {$M$};
%line
    \path [arrowline]  (Y) to (X);
    \path [arrowline]  (Z) to (X);
    \path [arrowline]  (M) to (X);
    
    %draw inference net
    \path [dasharrowline]  (X) to[out=-20,in=-70, distance=0.5cm ] (M);
    \path [dasharrowline]  (X) to[out=20,in=-20, distance=0.5cm ] (Z);
    \path [dasharrowline]  (Y) to[out=45,in=145, distance=0.5cm ] (Z);
    \path [dasharrowline]  (M) to[out=135,in=45, distance=0.5cm ] (Z);
   
    %Plates
    \node[surround, inner sep = .5cm] (f_N) [fit = (Z)(X)(Y)(M) ] {};
\end{tikzpicture}

%% file: model.tex
\vspace{-2pt}
\section{The Causal Manipulation Augmented Model }
\label{sec:model}
\vspace{-2pt}
We propose a deep CAusal Manipulation Augmented model (deep CAMA), which takes into account the causal relationship for model design. We also design a fine-tuning algorithm to enable adaptive reasoning of deep CAMA for unseen manipulations on effect variables. The robustness can be further improved by training-time data augmentation, without sacrificing the generalization ability to unseen manipulations. Below we first present the deep CAMA for single modality data, and then present a generic deep CAMA for multimodality measurement data.

\subsection{Deep CAMA for single modality data}
\label{sec:single_M_CAMA}
The task of predicting $Y$ from $X$ covers a wide range of applications such as image/speech recognition and sentiment analysis. Normally a discriminative DNN takes $X$ as input and directly predicts (the distribution of) the target variable $Y$. Generative classifiers, on the other hand, build a generative model $Y \rightarrow X$, and use Bayes' rule for predicting $Y$ given $X$: $p(y | x) = p(y) p (x | y) / p(x)$.
%

% \begin{figure}[t]
% \centering
% \scalebox{0.8}{\input{tikz_model.tex}}
% \caption{Graphical presentation of proposed causally consistent deep generative model for single modal data. }
% %\vspace{-5pt}
% \label{fig:CAMA1}
% \end{figure}

%\begin{figure}[t]
\begin{wrapfigure}{r}{0.4\textwidth}
\vspace{-32pt}
    \centering
    \includegraphics[width=0.95\linewidth, bb= 0 0 340 330]{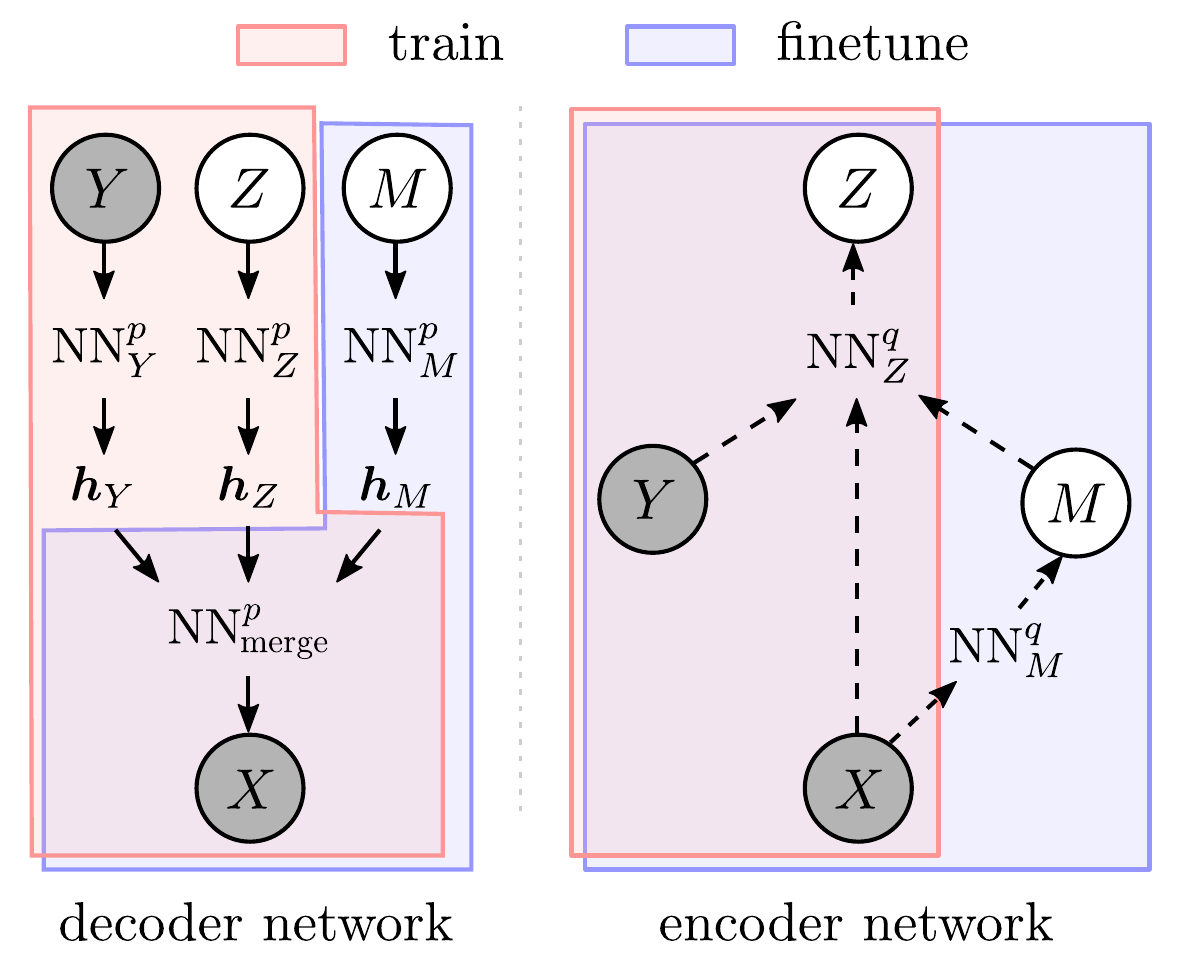}
    \caption{The network architecture. Shaded areas show the selective part for $do(m)$ training and the fine-tune method, respectively.}
    \vspace{10pt}
    \label{fig:network_architecture}
\end{wrapfigure}

We design deep CAMA (Figure \ref{fig:CAMA1}) following the causal graph in Figure \ref{fig:example}, which returns a factorized model:
\begin{equation}
    p_{\theta}(x, y, z, m) = p(m) p(z) p(y) p_{\theta}(x|y, z, m).
\end{equation}
Notice that we do not consider modelling dependencies between $Y$ and $M$ even when the causal relationship might exist and \ref{fig:example_confounding}) in the generation process of the training data (see Figures \ref{fig:example_y_cause_m}. By our definition of valid perturbation, $M$ is intervened on (i.e.,~$do(m)$), which blocks the influence from $Y$ to $M$, and the generation process of manipulated data reduces to the factorized case (Figure \ref{fig:example}).

For efficient posterior inference we use \emph{amortization} \citep{kingma2013auto, rezende2014stochastic,zhang2018advances} to define an inference network: 
\begin{equation}
    q_{\phi}(z, m | x, y) = q_{\phi_1}(z | x, y, m) q_{\phi_2}(m|x).
\end{equation}
Here the variational parameters are $\phi = \{ \phi_1, \phi_2 \}$, where $\phi_1$ is network parameter for the variational distribution $q_{\phi_1}(z | x, y, m)$, and $\phi_2$ is used for the $q_{\phi_2}(m | x)$ part. 
We assume in $q$ that given $X$, $Y$ does not contain further information about $M$. As a consequence, during inference, $Y$ and $M$ are conditionally independent given $X$, although it is not implied in the $p$ graphical model (i.e., the causal model). Therefore in $q_{\phi_2}(m | x)$ we only extract the information of $M$ from $X$, which, as we show later, allows deep CAMA to learn unseen manipulations.

The network architecture is presented in Figure \ref{fig:network_architecture}. For the $p$ model, the cause variables $Y$, $Z$ and $M$ are first transformed into feature vectors $h_Y, h_Z$ and $h_M$. Later, these features are merged together and then passed through another neural network to produce the distributional parameters of $p_{\theta}(x | y, z, m)$. For the approximate posterior $q$, two different networks are used to compute the distributional parameters of $q_{\phi_2}(m|x)$ and $q_{\phi_1}(z | x, y, m)$, respectively.

\paragraph{Model training} 
We describe the training procedure for two different scenarios. First, assume that during training, the model observes clean data $\mathcal{D}=\{ (x_n, y_n) \}$ only. In this case we set the manipulation variable $M$ to a null value, e.g.~$do(m = 0)$.  and train deep CAMA by maximizing the likelihood function $\log p(x, y | do(m=0))$ under training data. As there is no incoming edges to the manipulation variable $M$, the do-calculus can be reduced to the conditional distribution $p(x, y | do(m=0)) = p(x, y | m=0)$. 
Since this marginal distribution is intractable, we instead maximize the intervention evidence lower-bound (ELBO) with $do(m=0)$, i.e. $\max_{\theta, \phi} \mathbb{E}_{\mathcal{D}}[ \text{ELBO}(x, y, do(m=0) )]$, with the ELBO defined as (derived in Appenfix \ref{sec:appendix_derivation}):
% %%%%%%%%tmp%%%%%
% \begin{equation}
% \hspace{-0.5em}
% \begin{aligned}
% \text{ELBO}(x) 
%  := \mathbb{E}_{q_{\phi}(z |x)} \left[ \log \frac{p_{\theta}(x| z) p(z)}{q_{\phi}(z | x)} \right].\\
%  p(x,z) = p_{\theta}(x| z) p(z)\\
%  q_{\phi}(z | x)\\
% p(x,z, y, m ) = p_{\theta}(x| y, z, m) p(y)p(z)\\
% \end{aligned}
% \label{eq:intervention_elbo}
% \end{equation}
% %%%%%%%%%%%tmp%%%%%%%%%%%
%
\begin{equation}
\hspace{-0.5em}
\begin{aligned}
\text{ELBO}(x, y, do(m=0)) 
 := \mathbb{E}_{q_{\phi_1}(z |x, y, m=0)} \left[ \log \frac{p_{\theta}(x| y, z, m=0) p(y)p(z)}{q_{\phi_1}(z | x, y, m=0)} \right].
\end{aligned}
\label{eq:intervention_elbo}
\end{equation}
If manipulated data $\mathcal{D'}$ is available during training, then similar to data augmentation and adversarial training \citep{goodfellow:explaining2015,tramer:ensemble2017,madry2017towards}, we can augment the training data with such data. We still use the intervention ELBO (\ref{eq:intervention_elbo}) for clean data.
For the manipulated instances, we can either use the intervention ELBO with $do(m = m_0)$ when the manipulated data $\mathcal{D'}=\{(m_0(x), y) \}$ is generated by a known intervention $m_0$, or, as done in our experiments, infer the latent variable $M$ for unknown manipulations. This is achieved by maximizing the ELBO on the joint distribution $\log p(x, y)$ using manipulated data:
\begin{equation}
\text{ELBO}(x, y) := \mathbb{E}_{q_{\phi}(z, m | x, y)} \left[ \log \frac{p_{\theta}(x, y, z, m)}{q_{\phi}(z, m | x, y)} \right],
\label{eq:elbo}
\end{equation}
so the total loss function to be maximized is defined as
\begin{equation}
\begin{aligned}
    \mathcal{L}_{\text{aug}}(\theta, \phi) =  \lambda \mathbb{E}_{\mathcal{D}}[\text{ELBO}(x, y, do(m=0))] 
    + (1 - \lambda) \mathbb{E}_{\mathcal{D'}}[\text{ELBO}(x, y)].
\label{eq:training_data_aug}
\end{aligned}
\end{equation}

This training procedure only requires knowledge on whether the training data is clean (in such case we set $m=0$) or manipulated (potentially with unknown manipulation). In the manipulated case the model does not require explicit label for $M$ and performs inference on it instead. For test data, only the input X is given, and the labels for both $Y$ and $M$ are not available. 

Our causally consistent model effectively disentangles the latent representation: $Z$ models the unknown causes in the clean data, such as personal writing style; and $M$ models possible manipulations or interventions on the underlying factors, which the model should be robust to, such as shift, rotation, noise etc. 
From a causal perspective, the mechanism of generating $X$ from its causes is invariant to the interventions on $M$. 
Thus, in our model the functional relationships $Y \rightarrow X$ and $Z \rightarrow X$ remain intact even in the presence of manipulated data.
As a result, deep CAMA's can still generalize to unseen manipulations even after seeing lots of manipulated datapoints from other manipulations, in contrast to the behavior of discriminative DNNs as shown in Figure \ref{fig:disc_test}.

\paragraph{Prediction} 
We wish our model to be robust to an unseen intervention on test data $\mathcal{\tilde{D}}$, i.e.~$M$ is unknown at test-time. Here deep CAMA classifies an unseen test data $x^*$, using a Monte Carlo approximation to Bayes' rule with samples $m^u \sim q_{\phi_2}(m|x)$, $z_c^k \sim q_{\phi_1}(z| x^*, y_c, m^u)$:
\begin{align}
p(y^* | x^*) = \frac{p(x^* |y^*)p(y^*)}{p(x^*)} 
 \approx \text{softmax}_{c=1}^C\left[ \log \sum_{k=1}^{K} \frac{p_{\theta}(x| y, z_c^k, m^u) p(y_c)p(z)}{q_{\phi_1}(z_c^k | x^*, y_c, m^u)} \right].
\label{eq:classifiy_dom}
\end{align}

In experiments we use 1 sample of $m \sim q(m|x)$ and $K$ samples of $z \sim q(z|x, y, m)$ associated with each $m$ sample. Given the sampled $m$ and $z$ instances, we can compute the log-ratio term for each $y=c$ (as an approximation to $\log p(x, y=c)$), and apply softmax to compute Bayes' rule and obtain the predictive distribution. 

\paragraph{Test-time fine-tuning} 
Deep CAMA can also be adapted to the unseen manipulations presented in test data \emph{without labels}. From the causal graph, the conditional distributions $p(x | y)$ and $p(x | z)$ are invariant to the interventions on $X$ based on the modularity property.  
However, we would like to learn the manipulation mechanism $M \rightarrow X$, and, given that the number of possible interventions on $M$ might be infinity, the model may be underfitted for this functional relationship, given limited data. In this regard, fine-tuning on the current observation can be beneficial, thereby hopefully making deep CAMA more robust.  
As shown in Figure \ref{fig:network_architecture}, for the generative model, we only fine-tune the networks that are dependent only on $M$, i.e.~$\text{NN}_M^p$
by maximizing the ELBO of $\log p(x)$:
\begin{equation}
\text{ELBO}(x) := \log \left[ \sum_{c=1}^C \exp [\text{ELBO}(x, y_c)] \right]. %= \text{log-sum-exp}_{c=1}^C [ \text{ELBO}(x, y_c) ].
\label{eq:elbo_unsupervised}
\end{equation}
To reduce the possibly negative effect of fine-tuning to model generalization, we use a shallow network for $\text{NN}_{merge}^{p}$ and deep networks for $\text{NN}_M^p$, $\text{NN}_Y^p$ and $\text{NN}_Z^p$. 
We also fine-tune the network $\text{NN}_M^q$ for the approximate posterior $q$ since $M$ is involved in the inference of $Z$. In sum, in fine-tuning the selective part of the deep CAMA model is trained to maximize the following objective:
\begin{equation}
    \mathcal{L}_{\text{ft}}(\theta, \phi) = \alpha \mathbb{E}_{\mathcal{D}}[\text{ELBO}(x, y)] + (1 - \alpha) \mathbb{E}_{\mathcal{\tilde{D}}}[\text{ELBO}(x)].
\end{equation}
Note that the intervention ELBO can also be used for $\mathcal{D}$, in which we explore such option in some of the experiments. Importantly, there may exist infinitely many manipulations and it is impossible to train with all of them together. So by fine-tuning in a just-in-time manner, the model can be adapted to unseen manipulation at test time, which is confirmed in our experiments.%, the proposed deep CAMA model and the training methods are capable of improving the robustness of the generative classifier to unseen manipulations. 

The time complexity of CAMA training is in the same order of training a regular variational auto-encoder \citep{kingma2013auto,rezende2014stochastic}. For predictions, test-time fine-tuning requires a small amount of additional time, as only a small fraction of data is needed for fine-tuning, see Figure \ref{fig:FT_percentage} and Figure \ref{fig:FT_percentage_notdoM} in appendix.

\subsection{Deep CAMA for generic measurement data}
\label{sec:generic_CAMA}

We now discuss a generic version of deep CAMA to handle multimodality in measurement data. To predict the target variable $Y$ in a directed acyclic graph, only variables in the Markov blanket of $Y$ (shown in Figure \ref{fig:markove_blanket}) are needed. This includes the parents ($A$), children ($X$), and co-parents ($C$) of the target $Y$. Similar to the single modal case above, here a valid manipulation can only be independent mechanisms applied to $X$ or $C$ to ensure that both $Y$ and the relationship from $Y$ to $X$ remain intact.  

%\begin{wrapfigure}[8]{r}[0pt]{0.3\textwidth}
%\vspace{-20pt}
%\begin{figure}[t]
%\centering
%\begin{minipage}[]{0.45\linewidth}
%\centering
%\scalebox{0.58}{\input{tikz_markov_blanket.tex}}
%\caption{The Markov Blanket of target variable $Y$}
%\label{fig:markove_blanket}
%\end{wrapfigure}
%\end{minipage}
%
%\begin{wrapfigure}[9]{r}{0.3\textwidth}
%\vspace{-25pt}
%\begin{minipage}[]{0.45\linewidth}
%\centering
%\scalebox{0.58}{\input{tikz_model_Pa.tex}}
%\caption{Graphical presentation of deep CAMA for generic measurement  data. }
%\label{fig:general_CAMA}
%\end{minipage}
%\end{wrapfigure}

%\begin{wrapfigure}[8]{r}[0pt]{0.3\textwidth}
%\vspace{-20pt}
%\begin{figure}[t]
%\centering

%\end{wrapfigure}

\begin{minipage}[]{0.72\linewidth}
We design the generic deep CAMA (shown in Figure \ref{fig:general_CAMA}) following the causal process in Figure \ref{fig:markove_blanket}. Unlike discriminative DNNs where $A$, $C$ and $X$ are used together to predict $Y$ directly, we consider the full causal process and treat them separately.  Building on the deep CAMA for single modality data, we add the extra consideration of the parent and observed co-parent of $Y$, while modelling the latent unobserved cause in $Z$ and potential manipulations in $M$. We do not need to model manipulation on $C$ as they are out of the Markov Blanket of $Y$.  Thus, our model and the approximate inference network are defined as
\begin{equation}
\begin{aligned}
\hspace{-5pt}    p_{\theta}(x, y, z, m, a, c)
    = p(a) p(m) p(z) p(c) p_{\theta_1}(y | a) p_{\theta_2}(x|y,c, z, m),
\end{aligned}
\end{equation}
\vspace{-1em}
\begin{equation}
    q_{\phi}(z, m | x, y, a, c) = q_{\phi_1}(z | x, y, m, a, c) q_{\phi_2}(m|x).
\end{equation}
Training, fine-tuning and prediction proceed in the same way as in the single modality case (Section \ref{sec:single_M_CAMA}) with $do(m)$ operations and Monte Carlo approximations.
As we only fine-tune the networks that are dependent on $M$, similar reasoning indicates that the multimodality deep CAMA is robust to manipulations directly on the effect variable $X$.
\end{minipage}
\hfill
\begin{minipage}[]{0.25\linewidth}
\centering
%\vspace{3pt}
\scalebox{0.58}{\input{tikz_markov_blanket.tex}}
\captionof{figure}{The Markov Blanket of target variable $Y$}
\label{fig:markove_blanket}
\vspace{5pt}
\scalebox{0.52}{\input{tikz_model_Pa.tex}}
\captionof{figure}{Graphical presentation of deep CAMA for generic measurement  data. }
\label{fig:general_CAMA}
\end{minipage}

Our model is also robust to changes of $X$ caused by intervention on the co-parents $C$ by design. By our definition of valid manipulation, perturbing $C$ is valid as it only leads to the changes in $X$. If the underlying model from $Y$ and $C$ to $X$ remains the same, and the trained model learns $p(x | y, c)$ \emph{perfectly}, then our model is perfectly robust to such changes, due to Bayes' rule for prediction:
\begin{equation}
\begin{aligned}
    p(y |a, x, c ) = \frac{p(y| a) p(a) p(c) p(x|y,c)}{p(a)p(c) \int_y p(y|a) p(x|y,c) }
    = \frac{p(y| a)  p(x|y,c)}{ \int_y p(y|a) p(x|y,c) },
\end{aligned}
\label{eq:generic_CAMA_classify}
\end{equation}
and the manipulations on $C$ (thus changing $X$) do not affect the conditional distribution $p(x | y, c)$ in the generative classifier (Eq. \ref{eq:generic_CAMA_classify}). 
In contrast, discriminative DNNs concatenate $X$, $C$, $A$ together and map these variables to $Y$, and therefore it fails to make use of the invariant mechanisms.

\paragraph{Causal consistency in model design}
To build the deep CAMA model for measurement data (Figure \ref{fig:general_CAMA}), it requires a causally consistent specification of $C$, $X$, and $A$ variables in the graphical model. 
Thus, the causal view is crucial not only for valid manipulation definitions but also for model design. In this work, we assume that the causal relationship of the observed variables is given and use it to build the model, although in experiments we also empirically investigate the cases when this assumption is violated.
This is different from the line of causality research aiming at finding causal relations from observational data, where suitable assumptions are always needed \cite{spirtes2000causation,NSR_review18}. 
Orthogonal to our work, there are many methods for causal discovery (see, e.g.,  \cite{spirtes2000causation,glymour2019review}), for both observational data and interventional data, and remains an active research direction.

%% file: tikz_markov_blanket.tex
\pgfdeclarelayer{background}
\pgfdeclarelayer{foreground}
\pgfsetlayers{background,main,foreground}

\begin{tikzpicture}

\tikzstyle{scalarnode} = [circle, draw, fill=white!11,  
    text width=1.2em, text badly centered, inner sep=2.5pt]
\tikzstyle{arrowline} = [draw,color=black, -latex]
;
\tikzstyle{dasharrowline} = [draw,dashed, color=black, -latex]
;
ack,rounded corners=1mm]
%nodes
    \node [scalarnode] at (0,0) (Y)   {$Y$};
    \node [scalarnode] at (-2,0) (C1)   {$C_1$};
    \node [scalarnode] at (2,0) (C2)   {$C_2$};
    
    \node [scalarnode] at (-1,1.5) (A1)   {$A_1$};
\node [scalarnode] at (1,1.5) (A2)   {$A_2$};

    \node [scalarnode] at (-1,-1.5) (X1)   {$X_1$};
    \node [scalarnode] at (1,-1.5) (X2)   {$X_2$};
    
%line
    \path [arrowline]  (A1) to (Y);
    \path [arrowline]  (A2) to (Y);
    \path [arrowline]  (Y) to (X1);
    \path [arrowline]  (Y) to (X2);
    \path [arrowline]  (C1) to (X1);
    \path [arrowline]  (C2) to (X2);
\end{tikzpicture}

%% file: tikz_model_Pa.tex
\pgfdeclarelayer{background}
\pgfdeclarelayer{foreground}
\pgfsetlayers{background,main,foreground}

\begin{tikzpicture}

\tikzstyle{scalarnode} = [circle, draw, fill=white!11,  
    text width=1.2em, text badly centered, inner sep=2.5pt]
\tikzstyle{arrowline} = [draw,color=black, -latex]
;
\tikzstyle{dasharrowline} = [draw,dashed, color=black, -latex]
;
\tikzstyle{surround} = [thick,draw=black,rounded corners=1mm]
%nodes
    \node [scalarnode, fill=black!30] at (0,0) (X)   {$X$};
    \node [scalarnode, fill=black!30] at (-0.75-1.5, 1.5) (Y) {$Y$};
    \node [scalarnode, fill=black!30] at (-0.75, 1.5) (C) {$C$};
    \node [scalarnode] at (0.75, 1.5) (Z) {$Z$};
    \node [scalarnode] at (1.5+0.75, 1.5) (M) {$M$};
    \node [scalarnode, fill=black!30] at (-0.75-1.5, 3) (A) {$A$};
%line
    \path [arrowline]  (Y) to (X);
    \path [arrowline]  (C) to (X);
    \path [arrowline]  (Z) to (X);
    \path [arrowline]  (M) to (X);
    \path [arrowline]  (A) to (Y);
    
    %draw inference net
    \path [dasharrowline]  (X) to[out=-20,in=-70, distance=0.5cm ] (M);
    \path [dasharrowline]  (X) to[out=20,in=-40, distance=0.5cm ] (Z);
    \path [dasharrowline]  (Y) to[out=45,in=145, distance=0.5cm ] (Z);
    \path [dasharrowline]  (M) to[out=145,in=35, distance=0.5cm ] (Z);
    \path [dasharrowline]  (A) to[out=0,in=90, distance=0.5cm ] (Z);
    
    \path [dasharrowline]  (C) to[out=-30,in=210, distance=0.5cm ] (Z);
   
    %Plates
    \node[surround, inner sep = .5cm] (f_N) [fit = (Z)(X)(Y)(M)(A) ] {};
\end{tikzpicture}

%% file: exp.tex
\section{Experiments}
\label{sec:exp}
We evaluate the robustness of deep CAMA for image classification using both MNIST and a binary classification task derived from CIFAR-10 (Appendix \ref{app:CIFAR}). Furthermore, we demonstrate the behaviour of our generic deep CAMA for measurement data. We evaluate the performance of CAMA on both manipulations %such as shifting
and adverserial examples generated using the CleverHans package \citep{papernot2018cleverhans}. %More results with different DNN architectures and different manipulations are shown in the appendix.

\subsection{Robustness test on image classification with Deep CAMA}
We first demonstrate the robustness of our model to vertical (Ver) and horizontal (Hor) shifts. %Details such as network architectures are presented in the appendix. 

\begin{figure*}[t]
\vspace{-40pt}
\subfigure[Finetune Ver test Ver]{
\includegraphics[width=0.3 \textwidth, bb= 0 0 430 430 ]{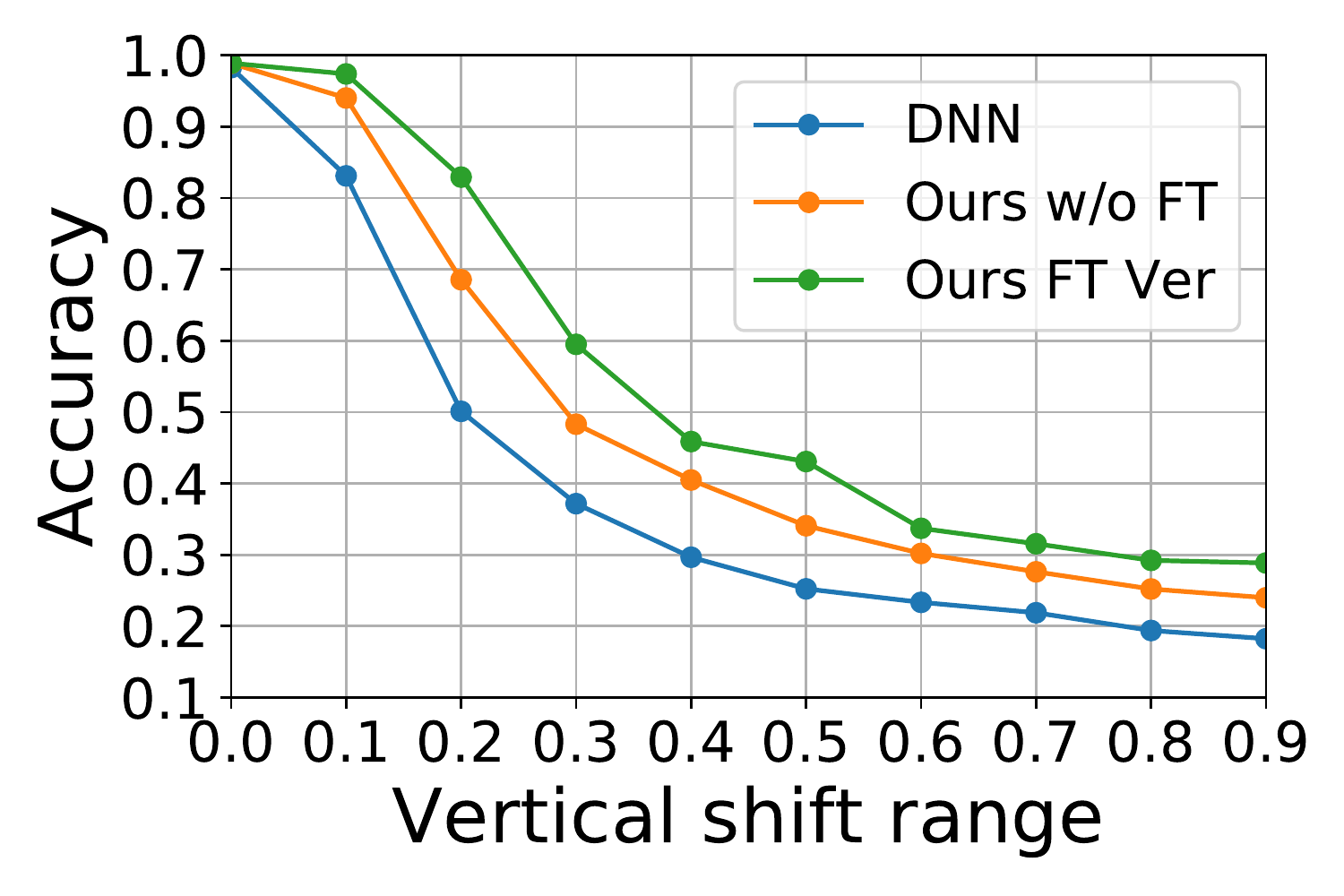}
\label{fig:VTVT}
}
\subfigure[Finetune Hor test Ver]{
\includegraphics[width=0.3 \textwidth, bb= 0 0 430 430]{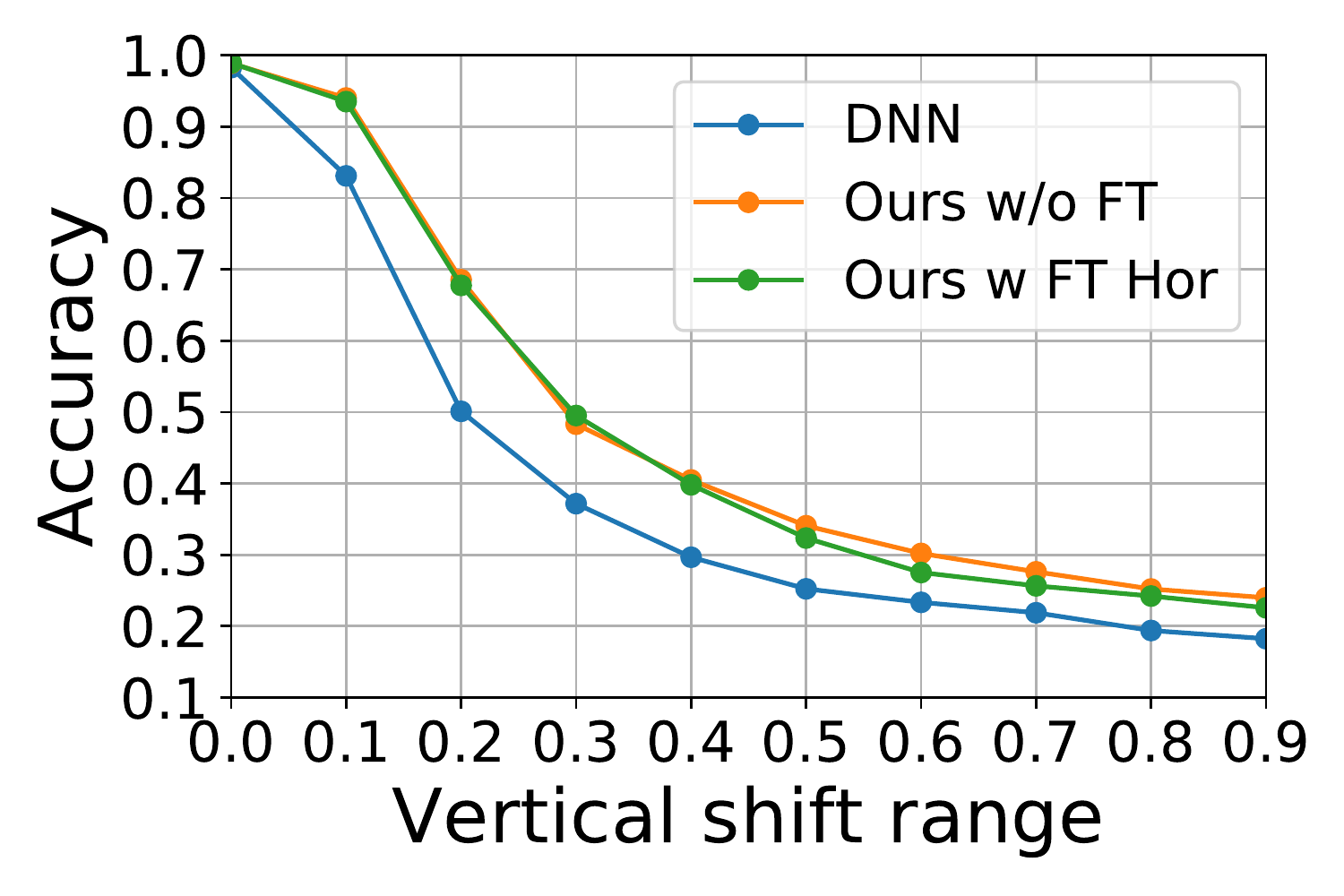}
\label{fig:HTVT}
}
\subfigure[Finetune Both test Ver]{
\includegraphics[width=0.3 \textwidth, bb= 0 0 430 430]{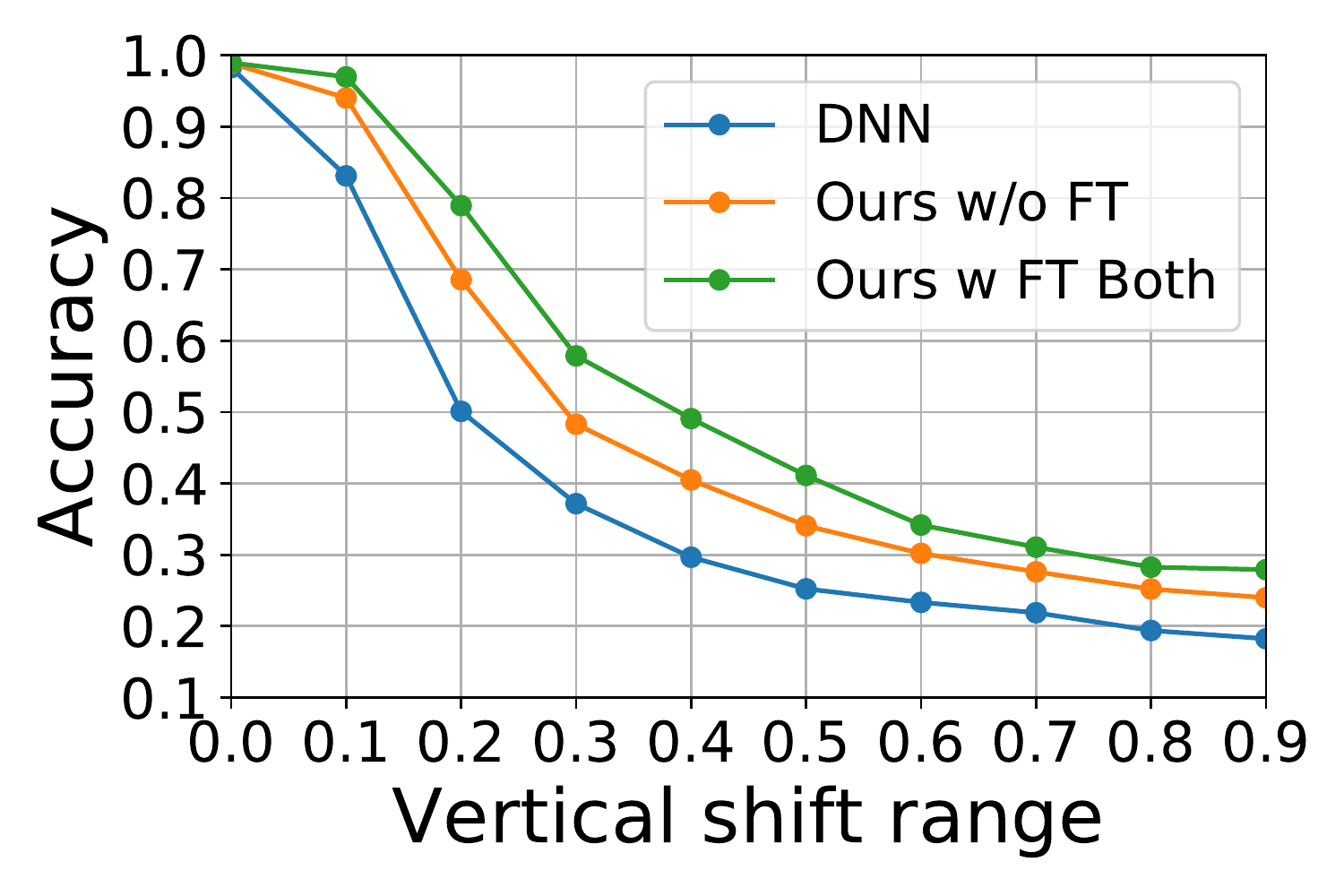}
\label{fig:BothVT}
}
\vspace{-2pt}
\caption{Model robustness results on vertical shifts. (See Appendix \ref{sec:app_MNIST} for results on horizontal shifts)}
\vspace{-10pt}
\label{fig:train_clean_mnist}
\end{figure*}

%\vspace{-5pt}
\paragraph{Training with clean MNIST} Figure \ref{fig:train_clean_mnist} shows the robustness results on vertical shifts for deep CAMA trained on clean data only.\footnote{Results on horizontal shifts are presented in Appendix \ref{sec:app_MNIST}, and the conclusions there are similar.}  Deep CAMA without fine-tuning (orange lines) perform slightly better than a DNN (blue lines). 
%\cz{In this case,  CAMA without fine-tuning can be viewed as such a baseline trained with VAE method.} 
The advantage of deep CAMA is clear when fine-tuning is used at test time (green lines): fine-tuning on manipulated test data with the same shift clearly improves the robustness of the network (Figure \ref{fig:VTVT}). 
Furthermore Figure \ref{fig:HTVT} shows that deep CAMA generalizes to \emph{unseen} vertical shifts after fine-tuning with horizontal shifts. Lastly, Figure \ref{fig:BothVT} shows the improved robustness of our model when both types of manipulation are used for fine-tuning. 
All these results indicate that our model is capable of learning manipulations in an unsupervised manner, without deteriorating the generalization ability to unseen manipulations. Comparisons to other deep generative model baselines are presented in the appendix \ref{sec:morebaseline}, and the results show the advantage of CAMA (especially with fine-tuning). 

%
%%%%%%%%exp figure%%%%
\begin{figure*}
\vspace{-15pt}
\centering
\begin{minipage}[]{0.36\textwidth}
    \centering
    \vspace{-18pt}
    \includegraphics[width=1\textwidth, bb= 0 0 430 430 ]{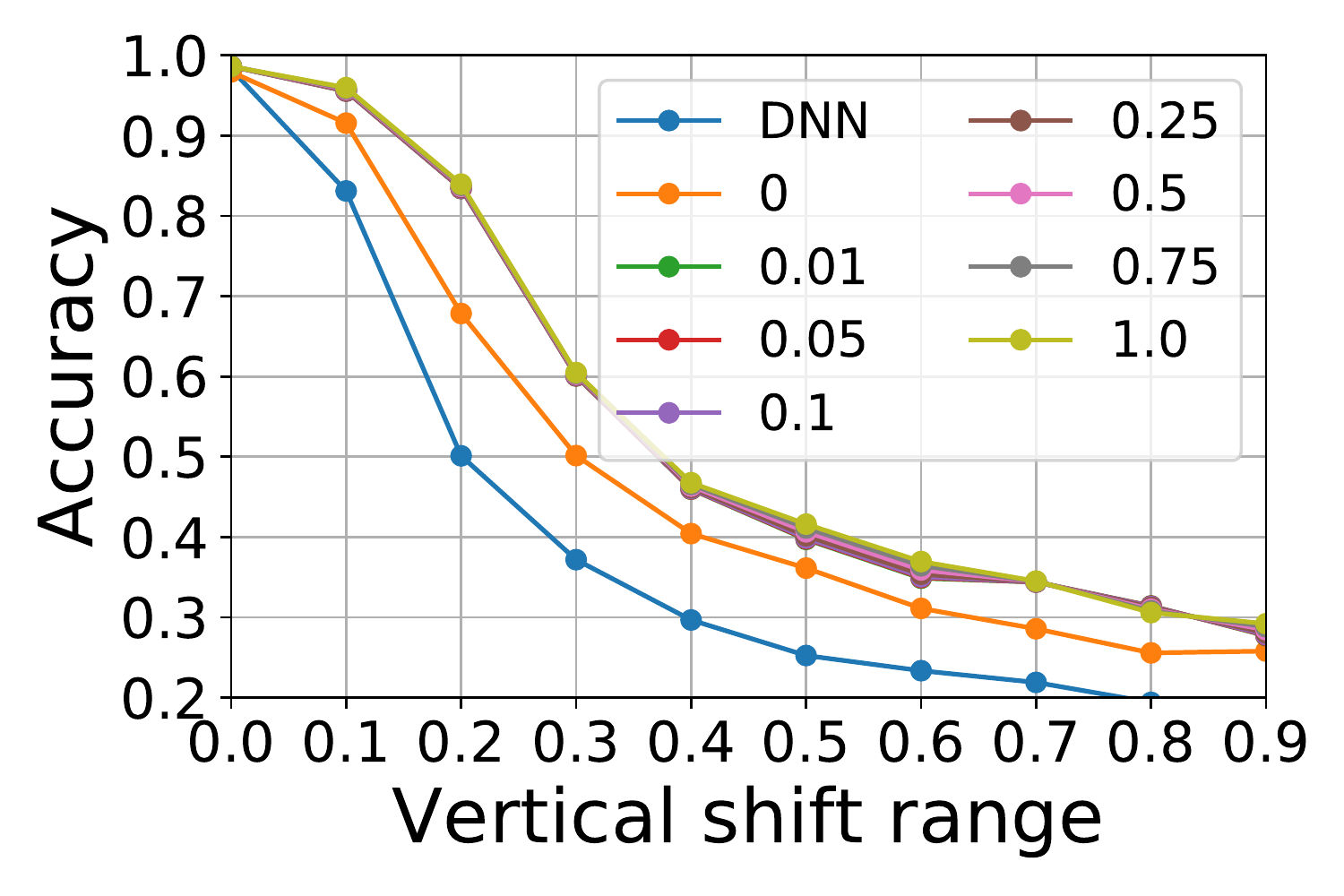}
    \caption{Performance regarding different percentages of test data used for fine-tuning manipulation}
    \label{fig:FT_percentage}
%\end{figure}
\end{minipage}
\hfill
\begin{minipage}[]{0.6\textwidth}
%\begin{figure}
    \centering
    \subfigure[Vertically shifted data]{
        \includegraphics[width=0.40\textwidth, bb= 0 0 300 300]{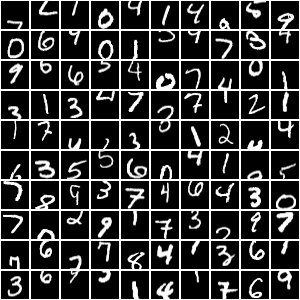}
        \label{fig:disentangle_1}
    }
    \hspace{0.1em}
        \subfigure[do(m=0) with $z$ \& $y$ from the vertical shifted data]{
        \includegraphics[width=0.40\textwidth, bb= 0 0 300 300]{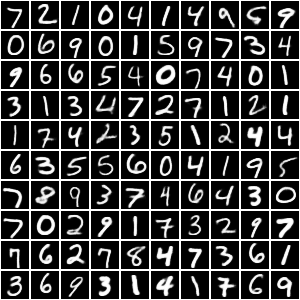}
        \label{fig:disentangle_2}
    }
    \vspace{-4pt}
    \caption{Visualization of the disentangled representation.}
    \vspace{-5pt}
    \label{fig:disentangle}
\end{minipage}
\vspace{-20pt}
\end{figure*}

%\vspace{-5pt}
\paragraph{Training with augmented MNIST} We explore the setting where the training data is augmented with manipulated data.
As discussed in Section \ref{sec:single_M_CAMA}, here deep CAMA naturally learns disentangled representation due to its causal reasoning capacity. Indeed this is confirmed by Figure \ref{fig:disentangle}, where panel \ref{fig:disentangle_2} shows the reconstructions of manipulated data from panel \ref{fig:disentangle_1} with $do(m=0)$. In this case the model keeps the identity of the digits but moves them to the center of the image. Recall that $do(m=0)$ corresponds to clean data which contains centered digits. This shows that deep CAMA can disentangle the intrinsic unknown style $Z$ and the shifting manipulation variable $M$.

%\begin{figure}
\begin{wrapfigure}[12]{r}{0.55\linewidth}
\vspace{-70pt}
    \centering
    \includegraphics[width=0.9\linewidth, bb= 0 0 860 450]{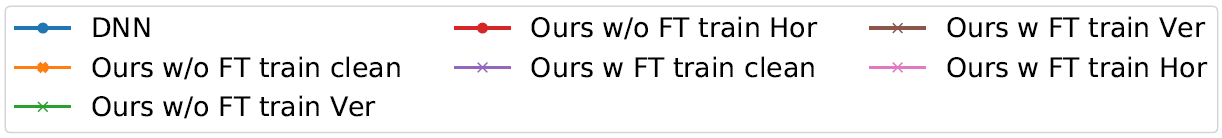}\vspace{-35pt}\\
\subfigure[Test Vertical shift]{
\includegraphics[width=0.45\linewidth, bb= 0 0 430 400]{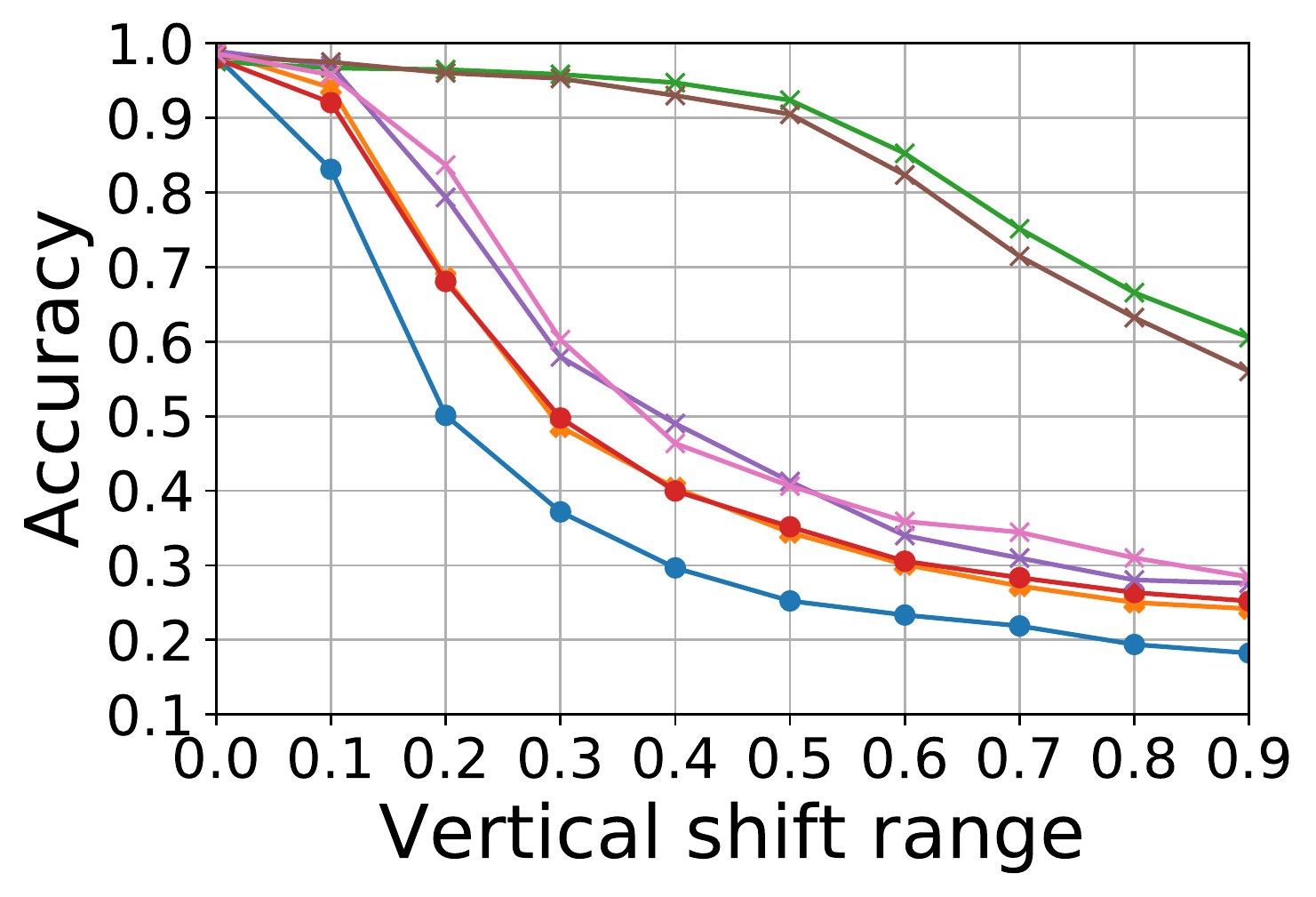}
\label{fig:our_VT}
}
\subfigure[Test horizontal shift]{
\includegraphics[width=0.45\linewidth, bb= 0 0 430 400]{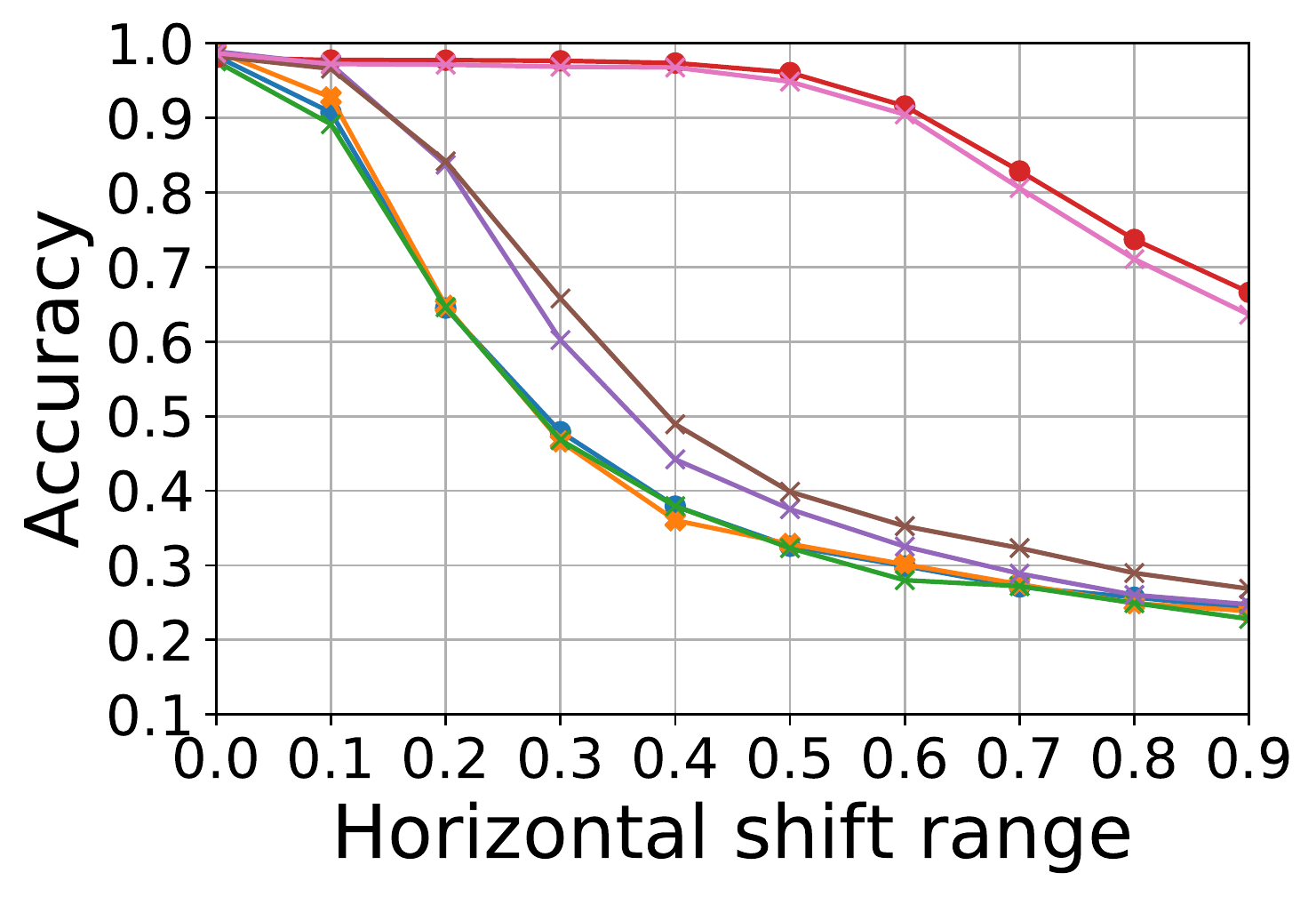}
\label{fig:our_HT}
}
\vspace{-5pt}
    \caption{Augmented MNIST robustness results.}
    \label{fig:ours_MNISt_shifts}
\end{wrapfigure}
%\vspace{-20pt}
%
We show the robustness results of deep CAMA with data augmentation (shift range 0.5) in Figure \ref{fig:ours_MNISt_shifts}. A comparison to results in Figures \ref{fig:disc_test} clearly shows the advantage of deep CAMA over disciminative DNNs: in addition to \emph{seen} perturbations in augmented data, deep CAMA is also robust to \emph{unseen} manipulation. Take the vertical shift test in panel \ref{fig:our_VT} for example. When vertically shifted data are augmented to the training set, the test performance without fine-tuning (green line) is significantly better. Further, fine-tuning (brown line) brings in even larger improvement for large scale shifts. 
On the other hand, deep CAMA maintains robustness on vertically shifted data when trained with horizontally shifted data. By contrast, training discriminative DNNs with one manipulation might even hurt its robustness to unseen manipulations (Figure \ref{fig:disc_test}). Therefore, our model does not overfit to a specific type of manipulations, at the same time further fine-tuning can improve the robustness against new manipulations (pink line). The same conclusion holds in panel \ref{fig:our_HT}.

We also quantify the amount of manipulated data required for fine-tuning in order to improve the robustness of deep CAMA models. As shown in Figure \ref{fig:FT_percentage}, even using $1\%$ of the manipulated data is sufficient to learn the vertical shift manipulation presented in the test set. 

%\vspace{-5pt}
\paragraph{Adversarial robustness on MNIST}

We further test deep CAMA's robustness to two adversarial attacks: fast gradient sign method (FGSM) \citep{goodfellow2014explaining} and projected gradient descent (PGD) \citep{madry2017towards}. Note that these attacks are specially developed for images with the small perturbation constraint. They are not guaranteed to be valid attacks by our definition, as the manipulation depends on $Y$, which has the risk of changing the ground-truth class label. Such risk has also been discussed in \citet{elsayed2018adversarial}.

% \begin{figure}[t]
\begin{wrapfigure}[9]{r}{0.55\linewidth}
\vspace{-100pt}
    \centering
    \includegraphics[width= 0.9\linewidth, bb= 0 0 860 460]{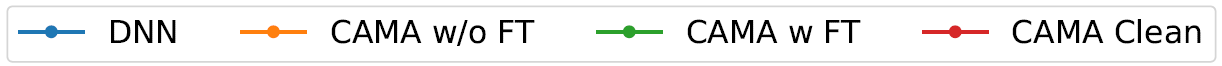}\vspace{-30pt}
    \includegraphics[width=0.45\linewidth, bb= 0 0 410 400]{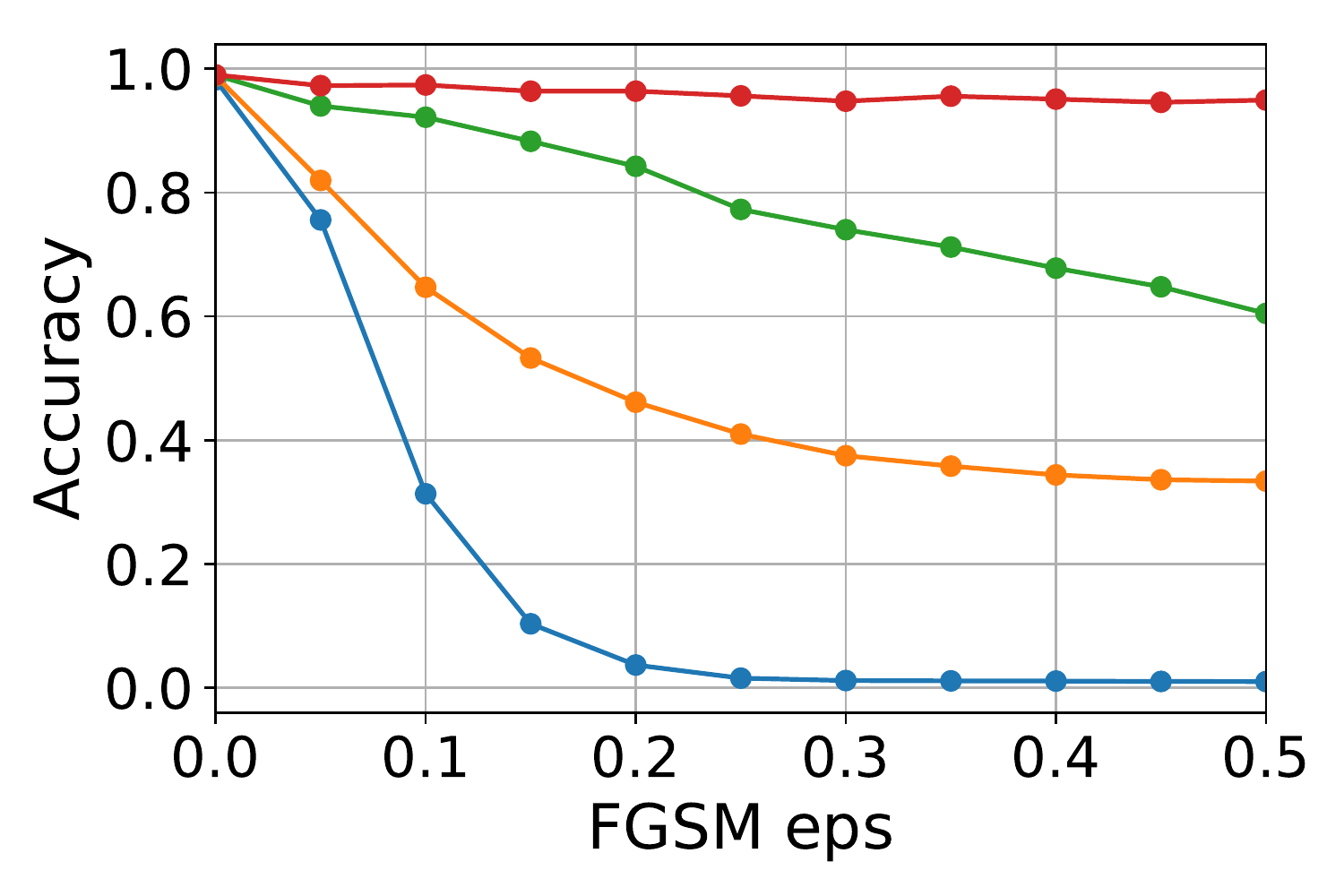}
    \includegraphics[width=0.45\linewidth, bb= 0 0 410 400]{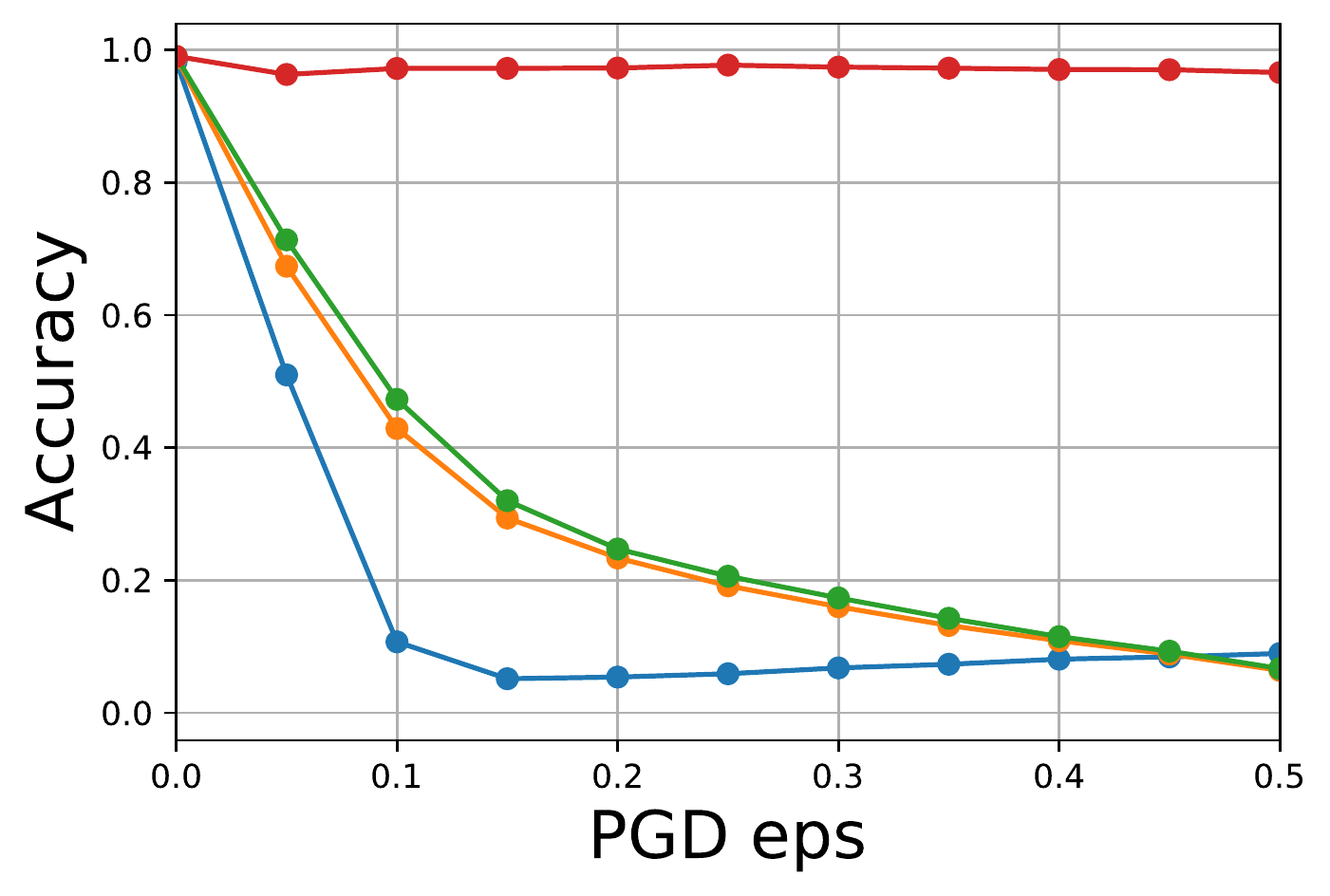}
       \vspace{-2pt}
    \caption{Test accuracy on MNIST adversarial examples.}
    \label{fig:MNIST_attack}
\end{wrapfigure}
%\vspace{-15pt}
%\end{figure}

Figure \ref{fig:MNIST_attack} show the results on models trained on clean images only.
Deep CAMA is significantly more robust to both attacks than the DNN, and with fine-tuning, deep CAMA shows additional $20\%- 40\%$ accuracy increase. The clean data test accuracy after fine-tuning remains the same thanks to the causal consistent model design. Comparisons to more baselines can be found in appendix \ref{sec:morebaseline}.

In Appendix \ref{app:CIFAR} we also perform adversarial robustness tests on a natural image binary classification task derived from CIFAR-10. Again deep CAMA out-performs a discriminative CNN even without fine-tuning; also fine-tuning provides additional advantages without deteriorating  the clean accuracy. 

%%%%%%%% CIFAR-binary results to appendix %%%%%%%%%%

\subsection{Robustness test on measurement based data with generalized Deep CAMA}

Our causal view on valid manipulations allows us to test model robustness on generic measurement data. Since there is no real-world dataset with known underlying causal graph, we generate synthetic data following a causal process with non-linear causal relationships (see Appendix \ref{app:exp_settings}), and perform robustness tests therein. We use Gaussian variables for $A$, $C$ and $X$, and categorical variables for $Y$.

%\vspace{-5pt}
\paragraph{Shifting tests} 
\begin{wrapfigure}[9]{r}{0.55\linewidth}
\vspace{-18pt}
    \subfigure[Manipulate co-parents $C$]{  
    \includegraphics[width=0.48\linewidth,height = 2.5cm, bb= 0 0 430 300]{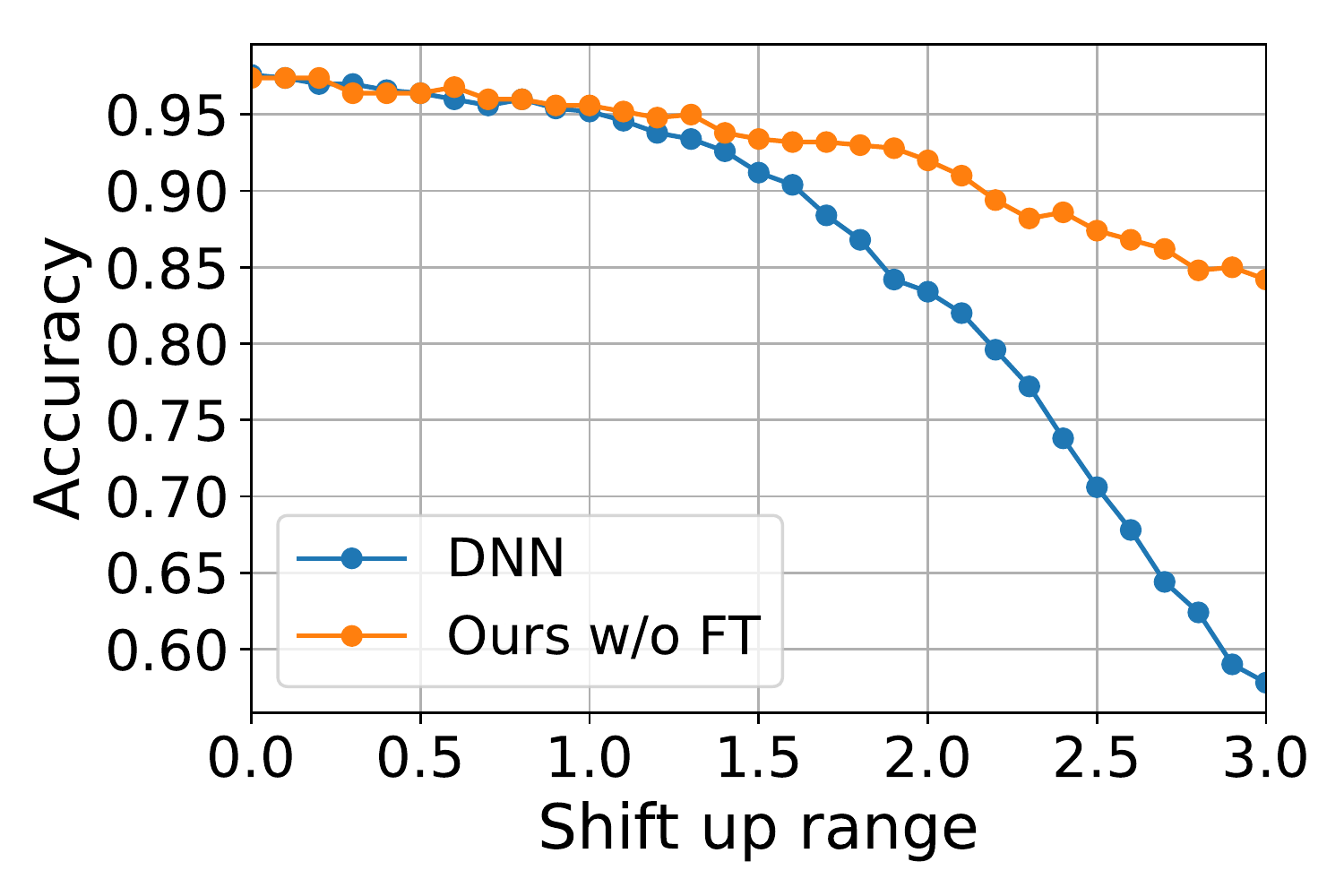}
    }
    \hspace{-10pt}
     \subfigure[Manipulate children $X$]{  \includegraphics[width=0.48\linewidth,height = 2.5cm, bb= 0 0 430 300]{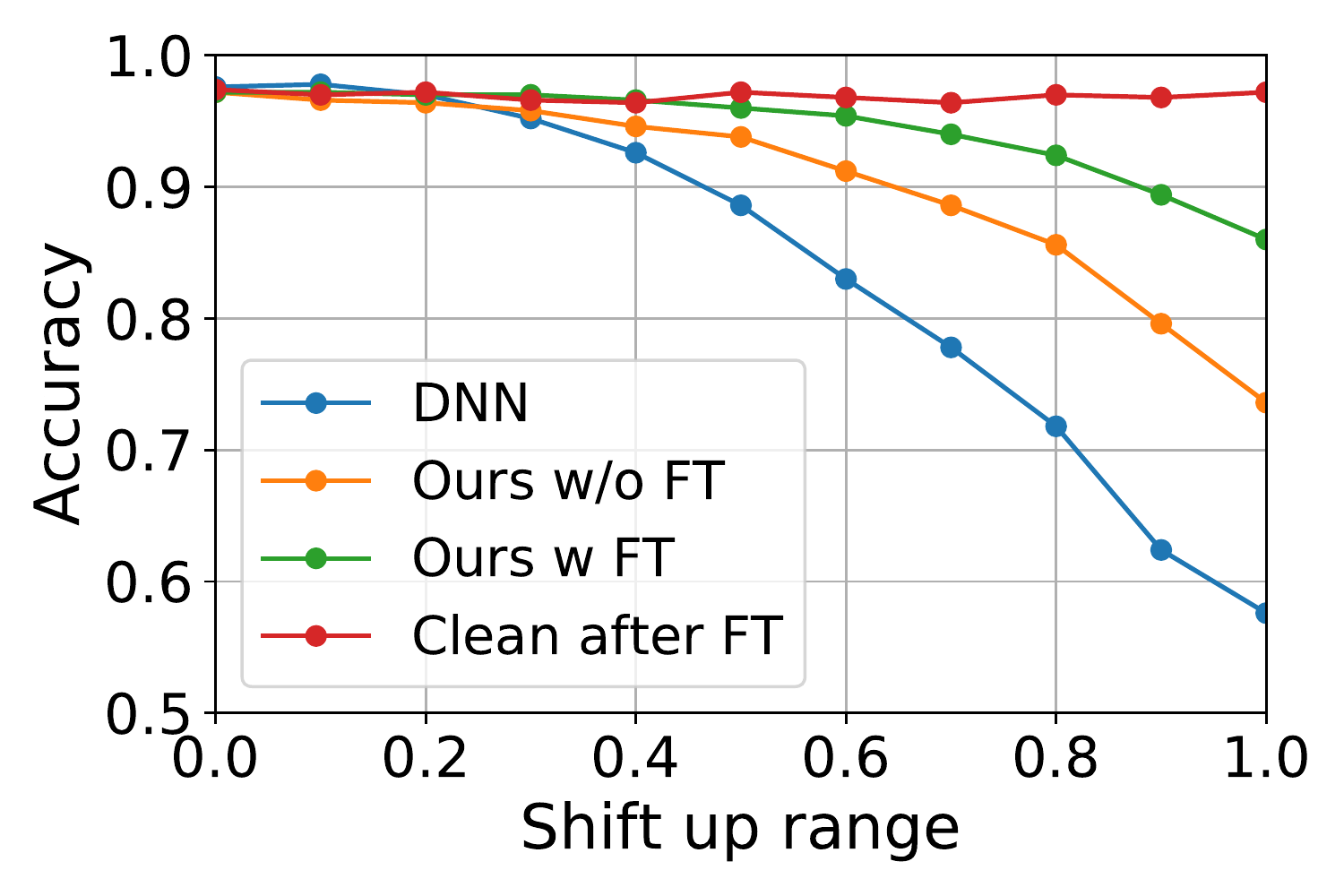}
    }
    \vspace{-5pt}
    \caption{Shift-up robustness results on measurement data.}
    %\vspace{-11pt}
    \label{fig:manipulate_measurement_up}
\end{wrapfigure}

We shift selected variables up- or down-scale, which resembles a type of noise in real-world data: different standards on subjective quantities, such as pain scale in diagnosis. We present the up-scale results below; the down-scale results in Appendix \ref{sec:app_measurement} are of similar behaviour.

The first test shifts the co-parents $C$ while keeping the relationship $C \rightarrow X$ static, resulting in a corresponding change in $X$.
Figure \ref{fig:manipulate_measurement_up}(a) shows the result: deep CAMA is significantly more robust. In particular, with increasing shift distortions, the accuracy of the DNN drops drastically. 
This corroborates our theory in Section \ref{sec:generic_CAMA} that manipulations in $C$ has little impact on deep CAMA's prediction. 

The second test shifts $X$ only, and the model only uses clean data for training. From Figure \ref{fig:manipulate_measurement_up}(b), deep CAMA is again much more robust even without fine-tuning (orange vs blue). This robustness is further improved by fine-tuning (in green) without negatively affecting clean test accuracies (in red). This confirms that fine-tuning learns the influence of $M$ without affecting the causal relationships between $Y$ and $Z$.

\paragraph{Adversarial robustness} 
\begin{wrapfigure}[9]{r}{0.55\linewidth}
\centering
\vspace{-30pt}
\subfigure[FGSM]{
    \includegraphics[width=0.48\linewidth, height=0.35\linewidth, bb= 0 0 430 300]{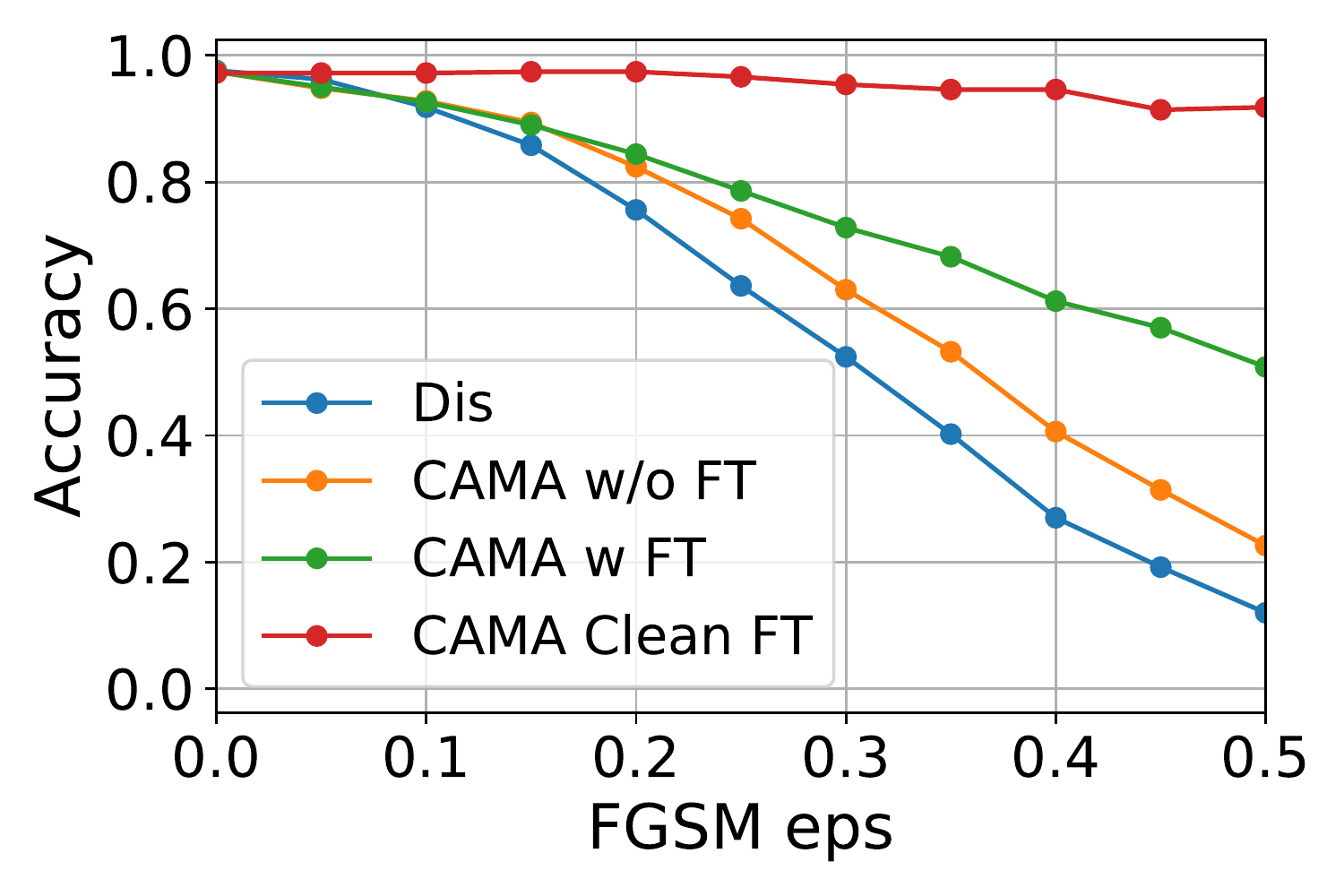}
    }
    \label{fig:measurement_FGSM}
    \hspace{-10pt}
\subfigure[PGD]{
    \includegraphics[width=0.48\linewidth, height=0.35\linewidth, bb= 0 0 430 300]{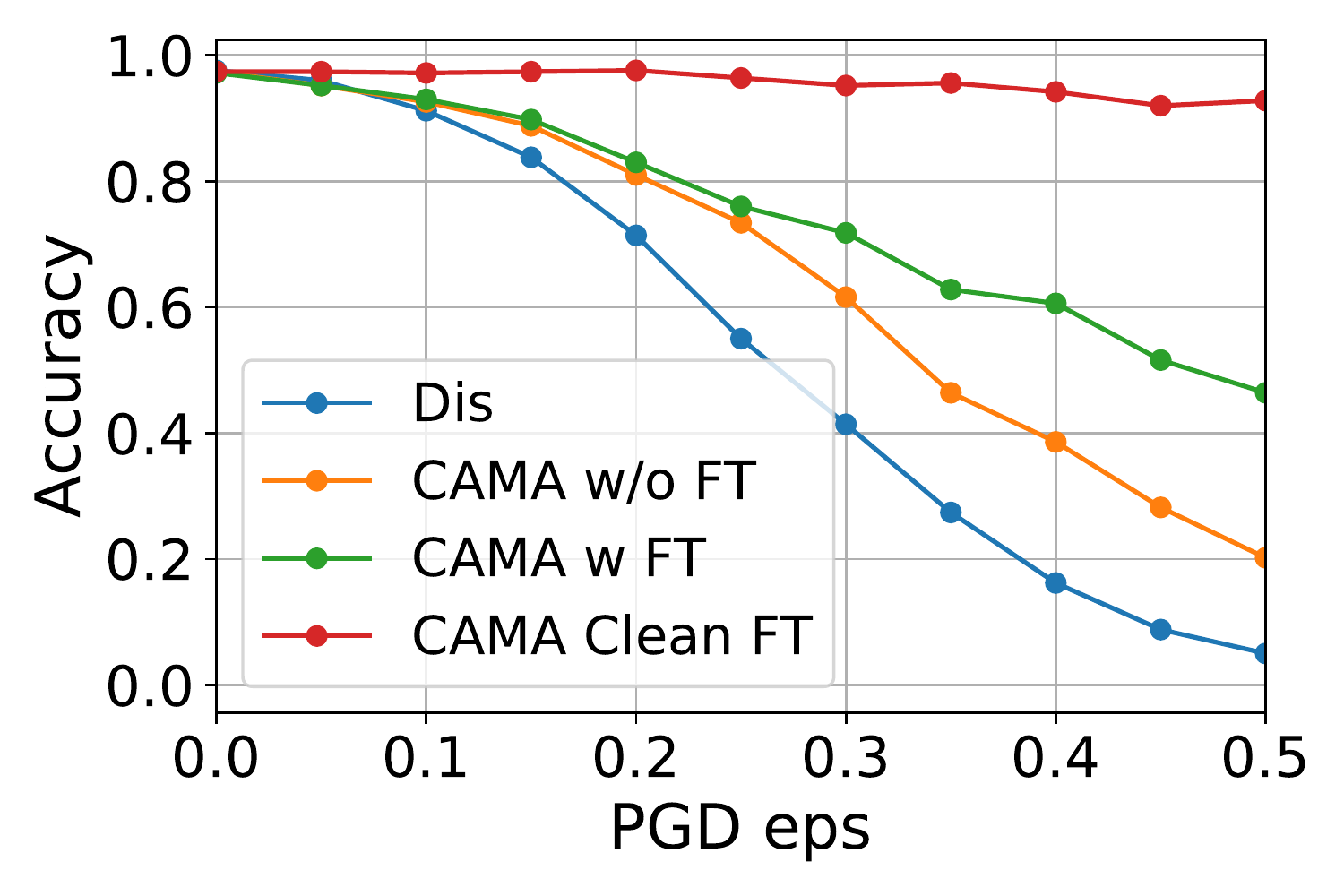}
    }
    \vspace{-5pt}
    \caption{Adv.~robustness results on measurement data.}
    \label{fig:measurement_attack}
\end{wrapfigure}
%\end{minipage}
%\end{figure*}
In this test we only allow attacks on the children $X$ and co-parents $C$ according to our definition of valid attacks. This applies to all the models in test. Figure \ref{fig:measurement_attack} shows again that deep CAMA are significantly more robust to adversarial attacks, and fine-tuning further improves robustness while keeping high accuracy on clean test examples.

\paragraph{Violated assumptions}
\begin{wrapfigure}[12]{r}{0.55\linewidth}
\vspace{-100pt}
    \centering
   \hspace{25pt} \includegraphics[width= 0.95\linewidth, bb= 0 0 680 350 ]{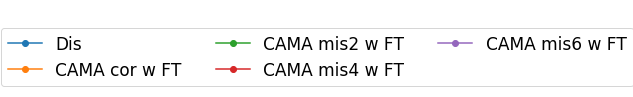}\vspace{-30pt}
    \includegraphics[width=0.45\linewidth, bb= 0 0 430 400 ]{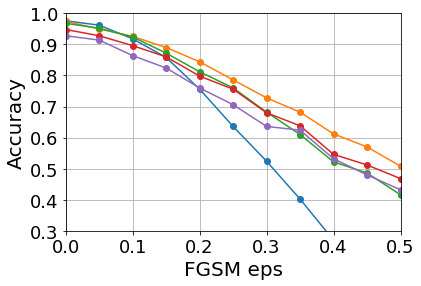}
    \includegraphics[width=0.45\linewidth, bb= 0 0 430 400 ]{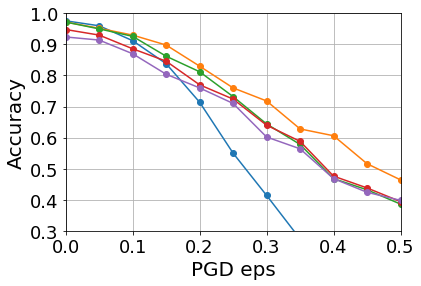}
       %\vspace{-5pt}
    \caption{Test accuracy on measurement data for CAMA with mis-specified model (``mis2'' denotes that 2 children nodes are mis-specified as co-parents). }
    \label{fig:Add_Child_to_CoP}
\end{wrapfigure}
So far we assume that the causal relationship among variables of interests are provided, either by domain experts or by running causal discovery algorithms. However, for both cases, there exists possibility that the provided causal graph is not perfect. We thus test the case where parts of deep CAMA's graphical model are mis-specified. It may happen when we have a wrong understanding of the data generating process, or  when a causal discovery algorithm  fail for multiple possible reasons.   
We use the same synthetic data as before, but to simulate this mis-specification setting, we intentionally use some children nodes in the ground truth causal graph as the co-parent nodes in deep CAMA's graphical model. Indeed this has negative impact on performance as shown in Figure \ref{fig:Add_Child_to_CoP}, however, deep CAMA remains to be more robust than the discriminative DNN when the mis-specification is not too severe. This again demonstrates the importance of causal consistency in model design. 
Additional results can be found in Appendix \ref{app:app_violate}. It also shows the importance of working closely with domain experts as well as careful evaluations of existing causal discovery algorithms \citep{tu2019neuropathic,glymour2019review}.

%% file: related.tex
\section{Related Work}
\label{sec:related}

\paragraph{Adversarial robustness}
Adversarial attacks can easily fool a discriminative DNN by adding imperceptible perturbations \citep{carlini:audio2018,alzantot2018nlp,carlini:bypass2017,szegedy2013intriguing,papernot:practical2017}. Adversarial training \citep{madry2017towards,tramer:ensemble2017} has shown some success in defending attacks; however, it requires knowledge of the adversary to present the perturbation to the model during training. Even so, a discriminative model after adversarial training is vulnerable to unseen manipulations. Meanwhile, existing theoretical works  \cite{dohmatob2018limitations,zhang2018efficient,fawzi2018adversarial} evaluate the robustness of a classifier trained on clean data and show there is no free lunch against attacks. However, such study does not necessary apply to our proposed model with test-time fine-tuning.
Deep generative modelling has been applied as a defence mechanism to adversarial attacks. One line of work considered de-noising adversarial examples before feeding these inputs to the discriminative classifier \citep{song:pixeldefend2018,samangouei:defensegan2018}. Another line of research revisited (deep) generative classifiers and provided evidence that they are more robust to adversarial attacks \citep{li2018generative,schott2018towards,lee2018simple}. 
Lastly, the Monte Carlo estimation techniques are also used in Bayesian neural networks (BNNs) which have also been shown to be more robust than their deterministic counterparts \citep{li2017dropout,feinman2017detecting}. However, \citet{li2018generative} shows that the advantage of generative classifiers over BNNs is due to the difference of the  generative/discriminative nature, rather than the usage of Monte Carlo estimates. Deep CAMA belongs to the class of generative classifiers, on top of \citet{li2018generative} we further demonstrates its improved robustness to unseen manipulations.

\paragraph{Causal learning} Causal inference has a long history in statistical research \citep{spirtes2000causation,pearl2009causality,peters2017elements,pearl2018book}, but to date, the causal view has not been widely incorporated to robust prediction under unseen manipulations. %
The most relevant work is in the field of transfer learning and domain adaption, where the difference in various domains are treated as either target shift or conditional shift from a causal perspective \citep{zhang2013domain, stojanov2019data,zhao2019learning, gong2016domain,ilse2019diva}. Extensions of the domain adaptation work also discuss robust predictors across different domains \citep{rothenhausler2018anchor,heinze2017conditional, arjovsky2019invariant}, in which the domain is specified either explicitly or though exemplar paired points. For example, invariant risk minimization (IRM) \cite{arjovsky2019invariant}, focusing on supervised learning with single modality data, proposes a regularizer in the loss function to encourage invariant predictions across different environments with given environments label in the training set. To make it adaptive and automated, domain adaptation has been viewed as problem of inference on graphical models, which provide a compact representation of the changeability of the data distribution and can be directly learned from data \cite{DM_Infer}. By contrast, our proposed method considers \emph{unseen} manipulations without relying on information of domain shifts. In addition, we address a more general problem -- interventions on a datapoint -- whereas in domain adaption interventions are considered on a dataset level. Another related area is causal feature selection \citep{aliferis2010local}, where causal discovery is applied first, and then features in the Markov Blanket of the prediction target are selected. 
We also note that CAMA's design is aligned with causal and anti-causal learning analyses \citep{scholkopf2012causal,kilbertus2018generalization}, in that CAMA models the causal mechanism $Y \rightarrow X$ and use Bayes' rule for anti-causal prediction.  
On the differences, CAMA is not limited to only two endogenous variables; rather it provides a generic design to handle latent causes that correspond to both intrinsic variations and data manipulations.

\paragraph{Disentangled representations}
Learning disentangled representations has become a trendy research topic in recent representation learning literature. Considerable effort went to developing training objectives, e.g.~$\beta$-VAE \citep{higgins2017beta} and other information theoretic approaches \citep{kim2018disentangling,chen2018isolating}. Additionally, different factorization structure in graphical model design has also been explored for disentanglement \citep{narayanaswamy2017learning,li2018disentangled}. The deep CAMA model is motivated by the causal process of data manipulations, which differs from the model used in \citet{narayanaswamy2017learning} in that the latent variables have different meanings. This difference is elaborated in Appendix \ref{app:compare_related_work}. Furthermore test-time fine-tuning allows deep CAMA to better adapt to unseen manipulations, which is shown to be useful for improving robustness.

%% file: discussion.tex
\section{Conclusion and Discussion}
\label{sec:discussion}
We provided a causal view on the robustness of neural networks, showing that the vulnerability of discriminative DNNs can be explained by the lack of causal reasoning. We defined valid attacks under this causal view, which are interventions of data though the causal factors which are not the target label or the ancestor of the target label. We further proposed a deep causal manipulation augmented model (deep CAMA), which follows the causal relationship in the model design, and  can be adapted to unseen manipulations at test time. Our model has demonstrated improved robustness, even without adversarial training. When manipulated data are available, our model's robustness increases for both seen and unseen manipulation. 

The ground truth causal graph is often complicated, but the CAMA graphical model simplifies it by grouping causes into different types, and treating each type as a single high dimensional variable. This is sufficient for our application at hand, but more fine-grind causal model may be needed for others. For example, one might prefer learning different type of manipulation separately, which may require working with a challenging causal graph. We leave the investigation in future work.

Our framework is generic, however, manipulations can change over time, and a robust model should adapt to these perturbations in a continuous manner. Our framework thus should be adapted to online learning or continual learning settings. In future work, we will explore the continual learning setting of deep CAMA where new manipulations come in a sequence. In addition, our method is designed for generic class-independent manipulations, and therefore a natural extension would consider class-dependent manipulations where $M$ is an effect of $Y$ or there is a confounder for $M$ and $Y$. Lastly, our design excludes gradient-based adversarial attacks which is dependent on both the target and the victim model. As such attacks are commonly adopted in machine learning, we would also like to extend our model to such scenarios.

\section*{Acknowledgement}
KZ would like to acknowledge the support by the United States Air Force under Contract No. FA8650-17-C-7715. CZ and YL would like to acknowledge Nathan Jones for his support with computing infrastructure; Tom Ellis and Luke Harries for feedback regarding the manuscript.

\section*{Broader Impact} %this can be in page 9
In this work, we provide a causal perspective on the robustness of deep learning, and propose a causal consistent deep generative model as an instance to improve the robustness of model regarding unseen manipulations. We view the robustness of AI solution under unseen manipulation as a key factor for many AI-aided decision making system to be trusted. While we do not intent to claim we have solved this problem perfectly (especially concerning large-scale and real-life applications), our work has shown great improvement over existing methods regarding the robustness towards unseen manipulations. We hope our research can inspire more solutions towards the final goal of AI safety.

%% file: appendix.tex
\appendix
\onecolumn

\section{Derivation Details}
\label{sec:appendix_derivation}

\subsection{The intervention ELBO}

When training with clean data $\mathcal{D}=\{(x_n, y_n) \}$, we set the manipulation variable $M$ to a null value, e.g. $do(m=0)$. In this case we would like to maximise the log-likelihood of the \emph{intervened} model, i.e.
$$\max_{\theta} \mathbb{E}_{\mathcal{D}}[\log p_{\theta}(x, y | do(m=0))].$$
This log-likelihood of the intervened model is defined by integrating out the unobserved latent variable $Z$ in the intervened joint distribution, and from do-calculus we have
\begin{equation}
\begin{aligned}
\log p_{\theta}(x, y | do(m=0)) &= \log \int p_{\theta}(x, y, z | do(m=0)) dz \\
&= \log \int p_{\theta}(x | y, z, m=0) p(y) p(z) dz.
\end{aligned}
\end{equation}
A variational lower-bound (or ELBO) of the log-likelihood uses a variational distribution $q(z | \cdot)$ 
\begin{equation}
\begin{aligned}
\log p_{\theta}(x, y | do(m=0)) 
&= \log \int p_{\theta}(x | y, z, m=0) p(y) p(z) \frac{q(z | \cdot)}{q(z | \cdot)} dz \\
&\geq \mathbb{E}_{q(z | \cdot)} \left[ \log \frac{p_{\theta}(x | y, z, m=0) p(y) p(z)}{q(z | \cdot)} \right].
\end{aligned}
\label{eq:appendix_elbo_general}
\end{equation}
The lower-bound holds for arbitrary $q(z| \cdot)$ as long as it is absolutely continuous w.r.t.~the posterior distribution $p_{\theta}(z | x, y, do(m=0))$ of the intervened model. Now recall the design of the inference network/variational distribution in the main text:
$$q_{\phi}(z, m | x, y) = q_{\phi_1}(z | x, y, m) q_{\phi_2}(m | x),$$
where $\phi_1$ and $\phi_2$ are the inference network parameters of the corresponding variational distributions. Performing an intervention $do(m=0)$ on this $q$ distribution gives
$$q_{\phi}(z | x, y, do(m=0)) = q_{\phi_1}(z | x, y, m=0).$$
Defining $q(z | \cdot) = q_{\phi_1}(z | x, y, do(m=0))$ and plugging-in it to eq.~(\ref{eq:appendix_elbo_general}) return the \emph{intervention} ELBO objective (\ref{eq:intervention_elbo}) presented in the main text.

\subsection{The ELBO for unlabelled test data}
The proposed fine-tuning method in the main text require optimising the marginal log-likelihood $\log p_{\theta}(x)$ for $x \sim \tilde{\mathcal{D}}$, which is clearly intractable. Instead of using a variational distribution for the unobserved class label $Y$, we consider the variational lower-bound of $\log p_{\theta}(x, y)$ for all possible $y = y_c$:
\begin{equation}
\begin{aligned}
\log p_{\theta}(x, y) &= \log \int p_{\theta}(x, y, z, m) dz dm \\
&= \log \int p_{\theta}(x, y, z, m) \frac{q_{\phi}(z, m | x, y)}{q_{\phi}(z, m | x, y)} dz dm \\
&\geq \mathbb{E}_{q_{\phi}(z, m | x, y)} \left[ \log \frac{p_{\theta}(x, y, z, m)}{q_{\phi}(z, m | x, y)} \right] := \text{ELBO}(x, y).
\end{aligned}
\end{equation}
Since both logarithm and exponent functions preserve monotonicity, and for all $y_c, c=1, ...,C$ we have $\log p_{\theta}(x, y_c) \geq \text{ELBO}(x, y_c)$, we have
$$\log p_{\theta}(x, y_c) \geq \text{ELBO}(x, y_c), \forall c \ \Rightarrow \ p_{\theta}(x, y_c) \geq \exp[\text{ELBO}(x, y_c)], \forall c $$
$$\ \Rightarrow \ \log p(x) = \log \left[ \sum_{c=1}^C p_{\theta}(x, y_c) \right] \geq \log \left[ \sum_{c=1}^C \exp[\text{ELBO}(x, y_c)] \right] := \text{ELBO}(x),$$
which justifies the ELBO objective (\ref{eq:elbo_unsupervised}) defined in the main text.

\section{Addition discussions on comparisons to \citet{narayanaswamy2017learning}}
\label{app:compare_related_work}
\citet{narayanaswamy2017learning} proposed a semi-supervised learning algorithm to learn disentangled representation for computer vision tasks. Their approach extends VAE with two latent variables $Y$ and $Z$ to model images $X$. They provide partially observed $Y$ which are ``interpretable''  depending on the computer vision application context. They achieve meaningful synthetic image generation by sample different latent variables in $Y$ and $Z$.  %This means in their model $Y$ is not limited to representing the prediction target; indeed their ``intrinsic face'' example the interpretable variables $Y$ include lightning and shading of the face image. Then semi-supervised learning algorithm achieves disentanglement by assuming the existence of a few supervision signal for the ``interpretable variables'' $Y$.

On the other hand, the proposed deep CAMA model focuses on a causal view for robustness of neural networks to unseen manipulations. Firstly, the latent variables have very different meanings. Apart from the fact that $Y$ is solely used to represent the prediction target, the $M$ and $Z$ variables are designed to separate the latent factors that can or cannot be \emph{artificially} intervened by the adversary. %Disentanglement of $M$ and $Z$ is achieved by training the model on interventional data (noisy data from valid manipulations).
Secondly, the fine-tuning algorithm provides a test-time adaptation scheme for the deep CAMA model (thus enabling adaptation to any unseen manipulations), which is different from \citet{narayanaswamy2017learning} where the trained model is directly applied in the testing time. Importantly, the fine-tuning method updates the model parameter in a selective manner, which is motivated by our analysis on the causal generation process of noisy data. It also allow the model to generate to unseen manipulation in test time. 

%Our model design can also to be applied beyond computer vision tasks. A generalized CAMA models all the variables in the Markov Blanket in a causal consistent manner to obtain robustness, which is clearly beyond \citet{narayanaswamy2017learning}.

Another important contribution of the paper is the generalization of deep CAMA to generic measurement data. In this case the causal graph of the data generation process plays a key role in model design, since now deep CAMA also models all the variables in the Markov blanket of $Y$, which is clearly beyond the factor model considered in \citet{narayanaswamy2017learning}. Our empirical study on the causal consistency of model design (see the ``violated assumptions'' experiments) clearly shows that a consistent design of the model to the underlying causal graph is key to both the robustness of the model and the efficiency of fine-tuning for unseen manipulations.

\section{Additional Results}
\subsection{Additional results of DNN behavior}
\paragraph{CNN}
We also performed experiments using different DNN network architectures. The convolution layers in CNN are designed to be robust to shifts. Thus, we test these vertical and horizontal shifts with a standard CNN architecture as used in \url{https://keras.io/examples/cifar10_cnn/}. 4 convolution layers are used in this architecture.

\begin{figure}[h]
    \centering
    \subfigure[Test Vertical shift]{
    \includegraphics[width = 0.4 \textwidth]{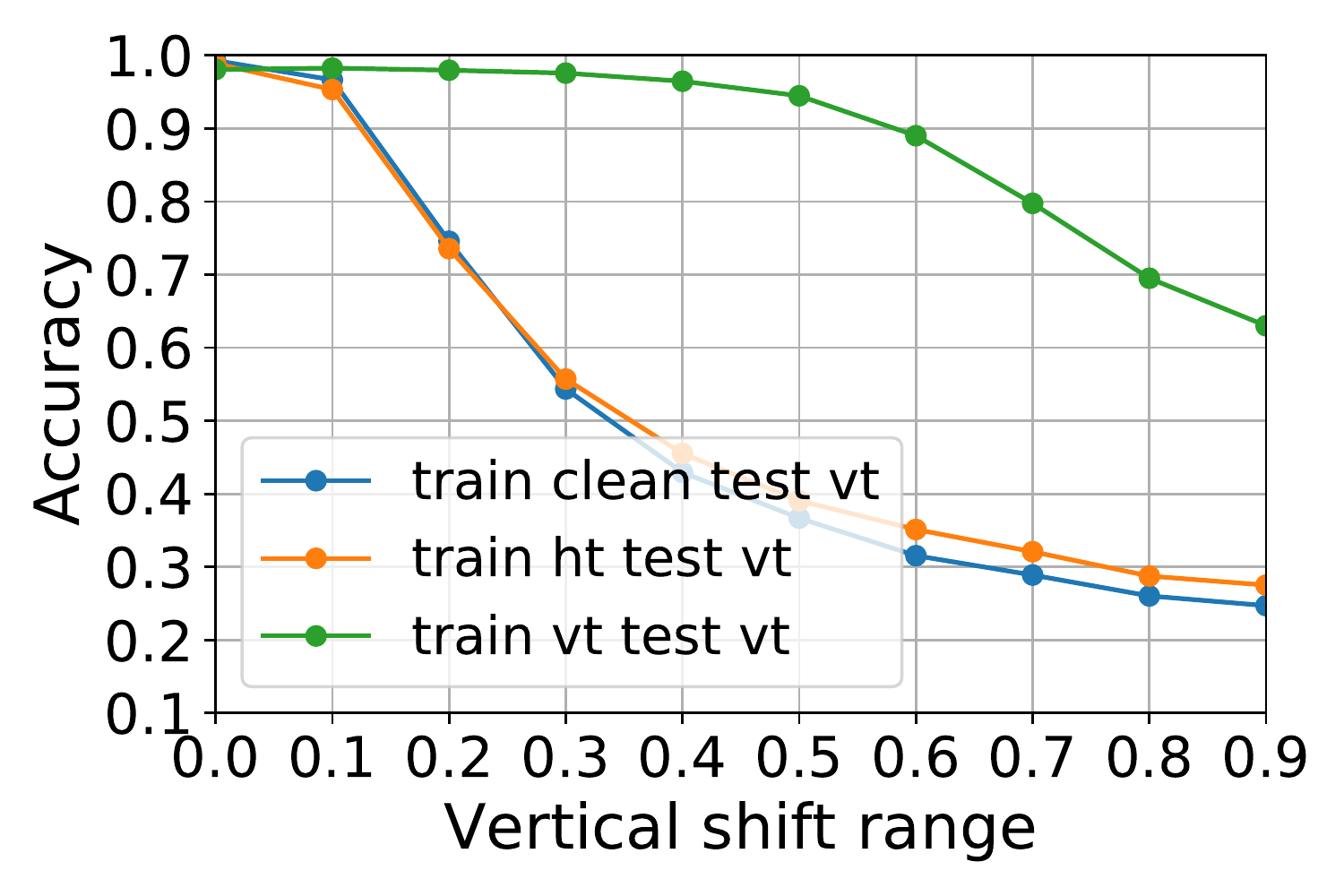}
    \label{fig:disc_vt_cnn}
    }
    \subfigure[Test Horizontal shift]{
    \includegraphics[width = 0.4 \textwidth]{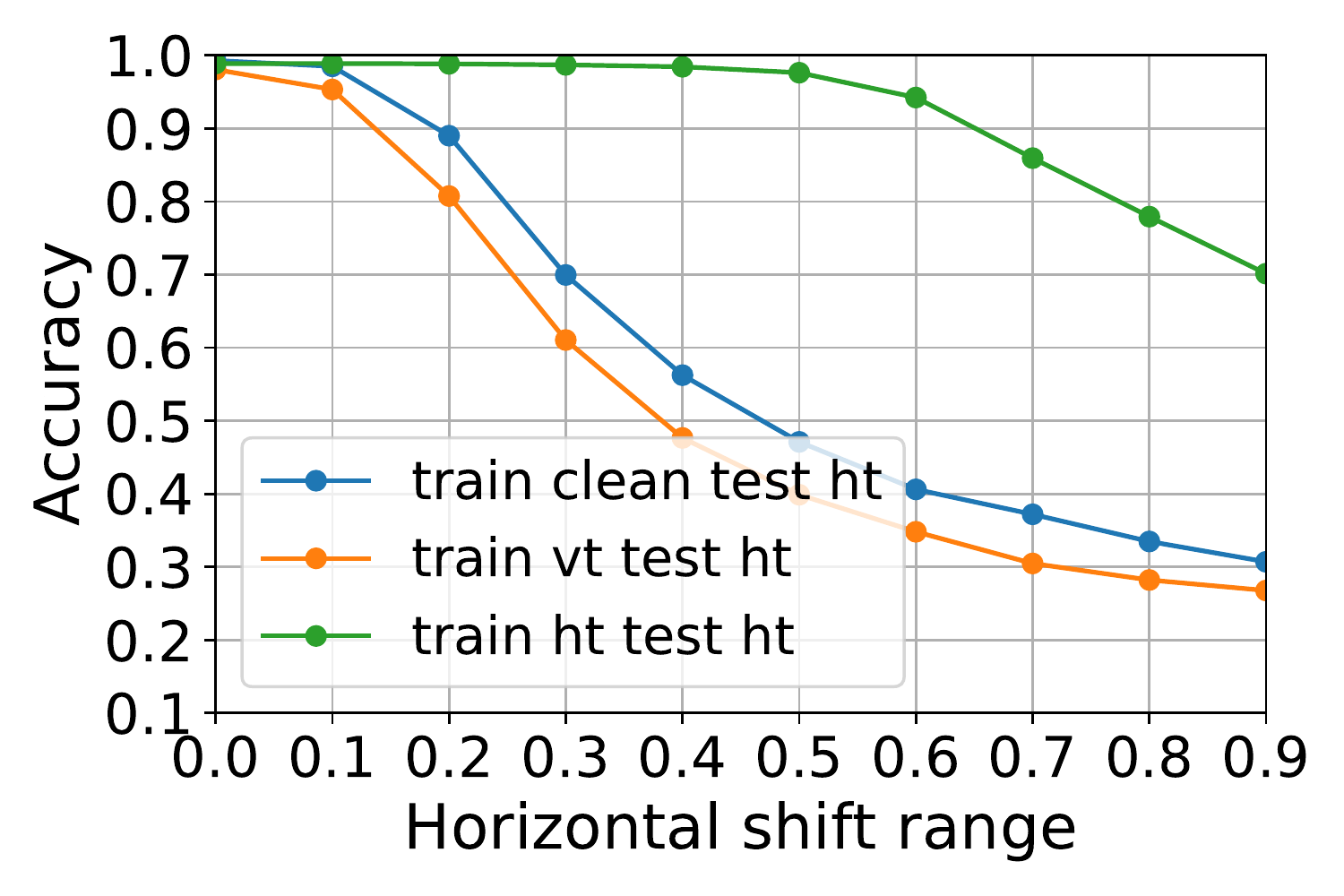}
    \label{fig:disc_ht_cnn}
    }
    \caption{
    Robustness results for DNNs against different manipulations on MNIST using CNN. Panels (a) and (b) show the accuracy on classifying noisy test data generated by shifting the digits vertically (vt) and horizontally (ht). It shows that data augmentation during training makes generalization to unseen shifts worse (orange versus blue lines).}
\label{fig:disc_test_cnn}
\end{figure}

Figure \ref{fig:disc_test_cnn} shows the performance against different shifts. We see that adding vertical shifts to the training data clearly harmed the robustness performances to unseen horizontal shifts as shown in \ref{fig:disc_ht_cnn}. Adding horizontal shifted images in training did not influences the performance on vertical shifts much. Thus, we see that using different architectures of DNN, even the one that are designed to be robust to these manipulations, lack of generalization ability to unseen data is a common problem. 

\begin{figure}[h]
    \centering
    \subfigure[Test Vertical shift]{
    \includegraphics[width = 0.4 \textwidth]{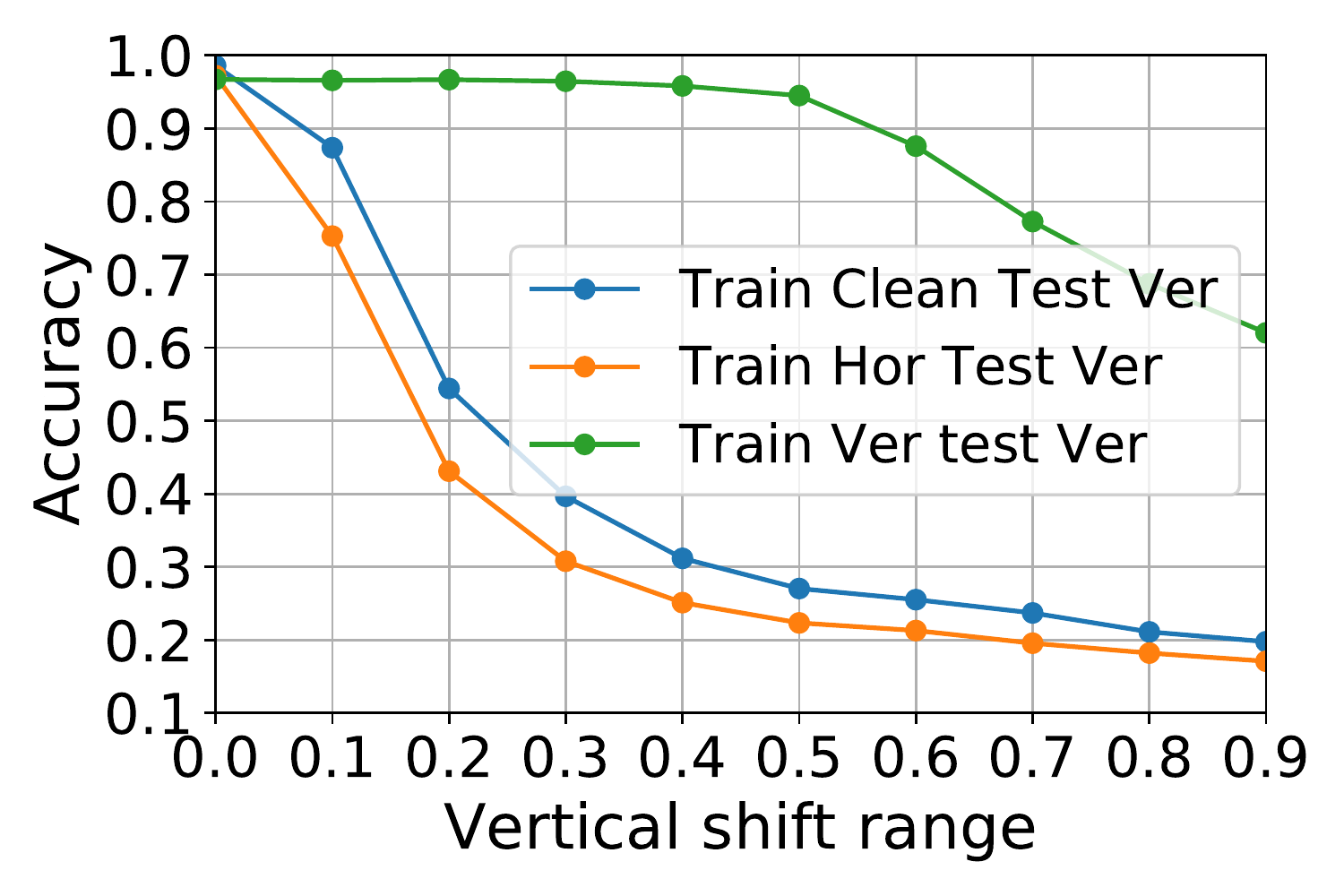}
    \label{fig:disc_vt_cnn}
    }
    \subfigure[Test Horizontal shift]{
    \includegraphics[width = 0.4 \textwidth]{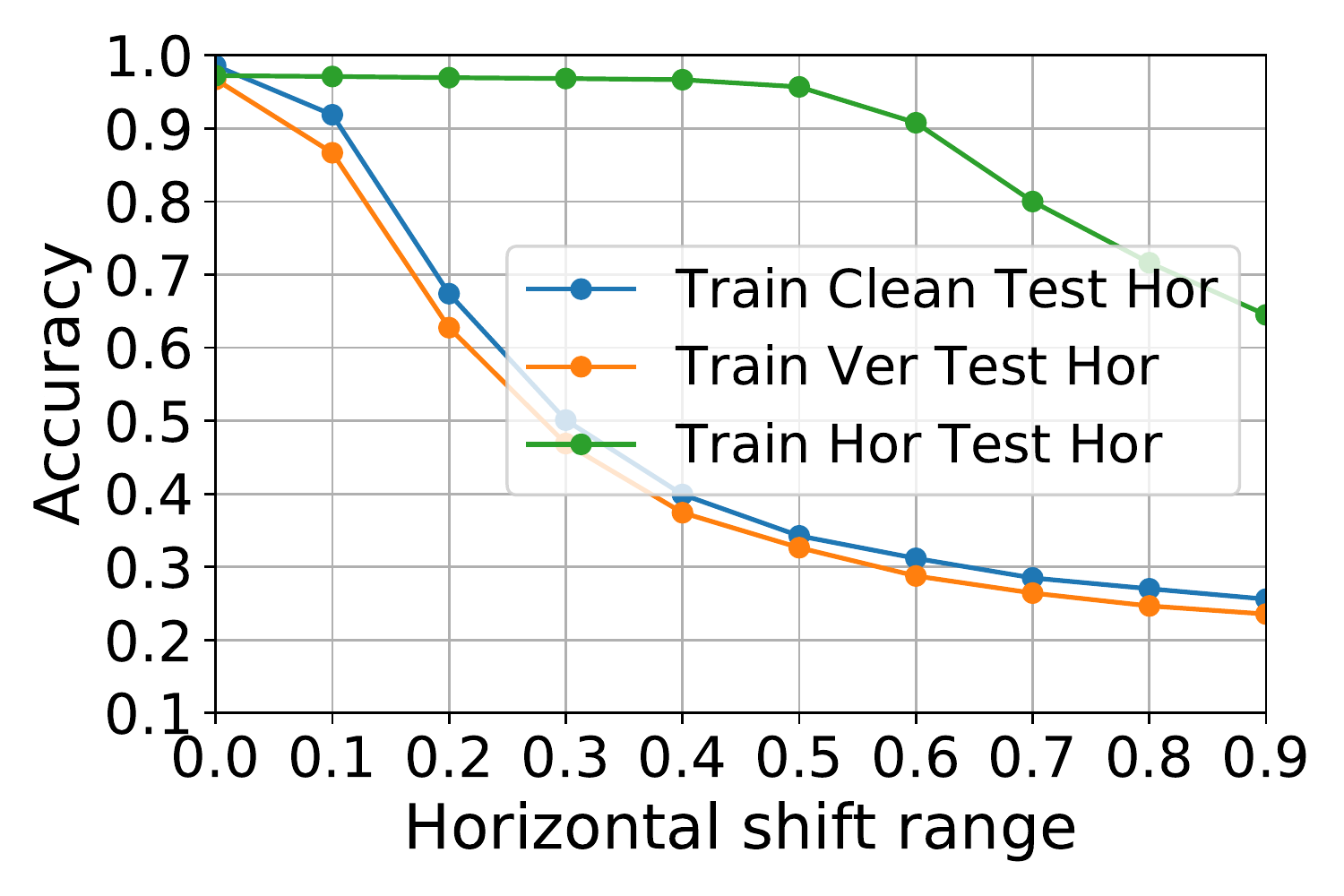}
    \label{fig:disc_ht_cnn}
    }
    \caption{
    Robustness results for DNNs against different manipulations on MNIST using a large MLP. Panels (a) and (b) show the accuracy on classifying noisy test data generated by shifting the digits vertically (vt) and horizontally (ht). It shows that data augmentation during training makes generalization to unseen shifts worse (orange versus blue lines).}
\label{fig:disc_test_larger_mlp}
\end{figure}
\paragraph{Enlarge Network Size}
Here we exam whether network capacity has any influence on the robustness performance to unseen manipulation. We use a wider network with [1024, 512, 512, 1024] units in each hidden layer instead of [512, 256, 126, 512] sized network in the paper. 
Figure \ref{fig:disc_test_larger_mlp} shows the robustness performance using this enlarged network. We observe the similar degree of over-fitting to the augmented data. The penalization ability shows no improvement by enlarging the network sizes.

\begin{figure}[h]
    \centering
    \subfigure[Test Vertical shift]{
    \includegraphics[width = 0.4 \textwidth]{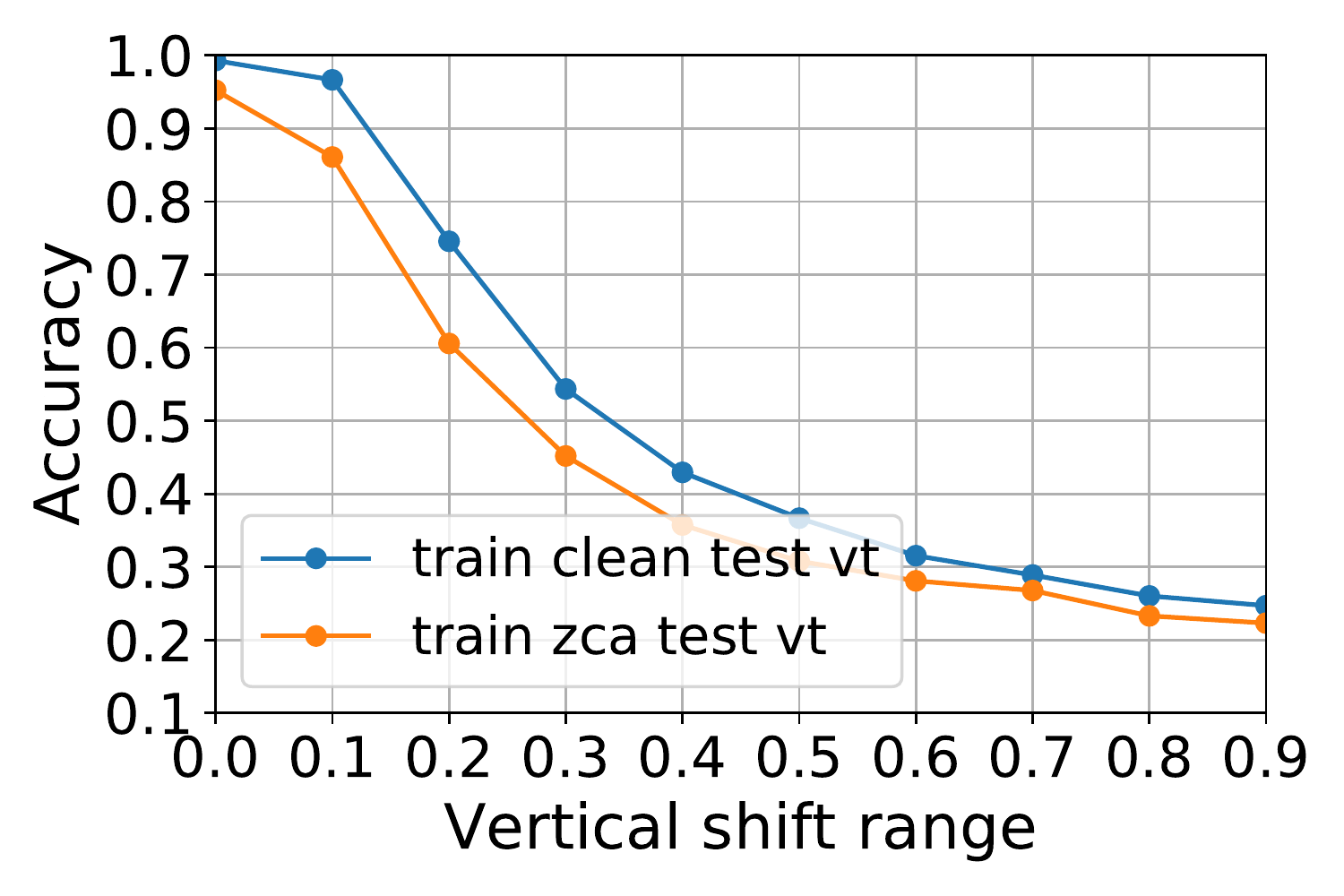}
    \label{fig:disc_vt_cnn_zca}
    }
    \subfigure[Test Horizontal shift]{
    \includegraphics[width = 0.4 \textwidth]{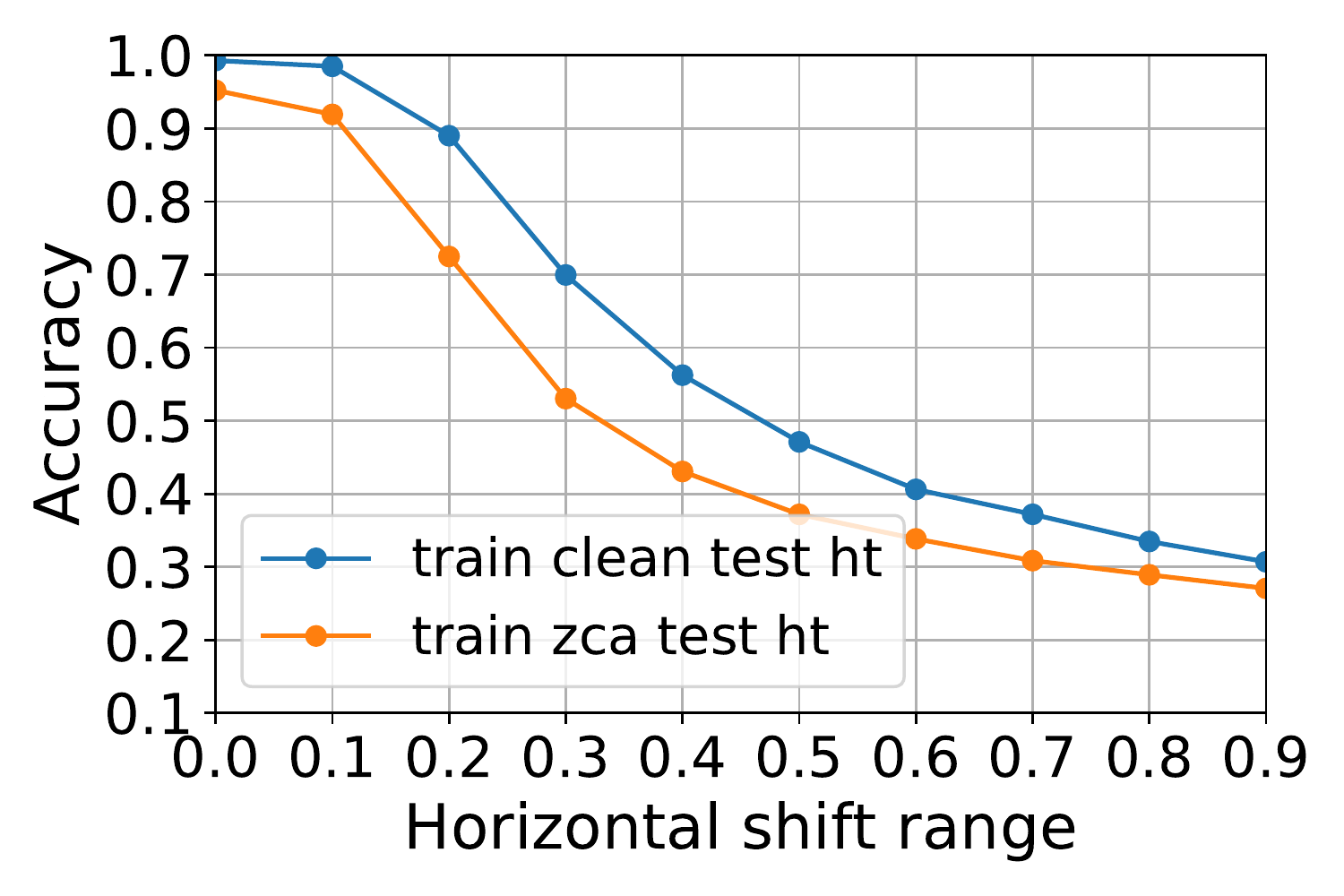}
    \label{fig:disc_ht_cnn_zca}
    }
    \caption{
    ZCA Whitening manipulation result. Figure shows the robustness results for DNNs against different manipulations on MNIST using CNN. The blue curve shows that result from training with clean data. The orange curve shows that result from training with zca whitening data added.}
\label{fig:disc_test_cnn_zca}
\end{figure}

\paragraph{ZCA Whitening Manipulation}
Our result does not limited to shifts, it generalizes to other manipulations. 
Figure \ref{fig:disc_test_cnn_zca} compare the result from training with clean images and training with ZCA whitening images added. 
We see that adding ZCA whitening images in training harm both robustness against vertical shift and horizontal shift. 

% \paragraph{Additional Figures}
% \label{app:additional_figures}
\subsection{Additional Results with MNIST experiment}
\label{sec:app_MNIST}
\begin{figure*}[t]
\subfigure[Finetune VT test HT]{
\includegraphics[width=0.3 \textwidth]{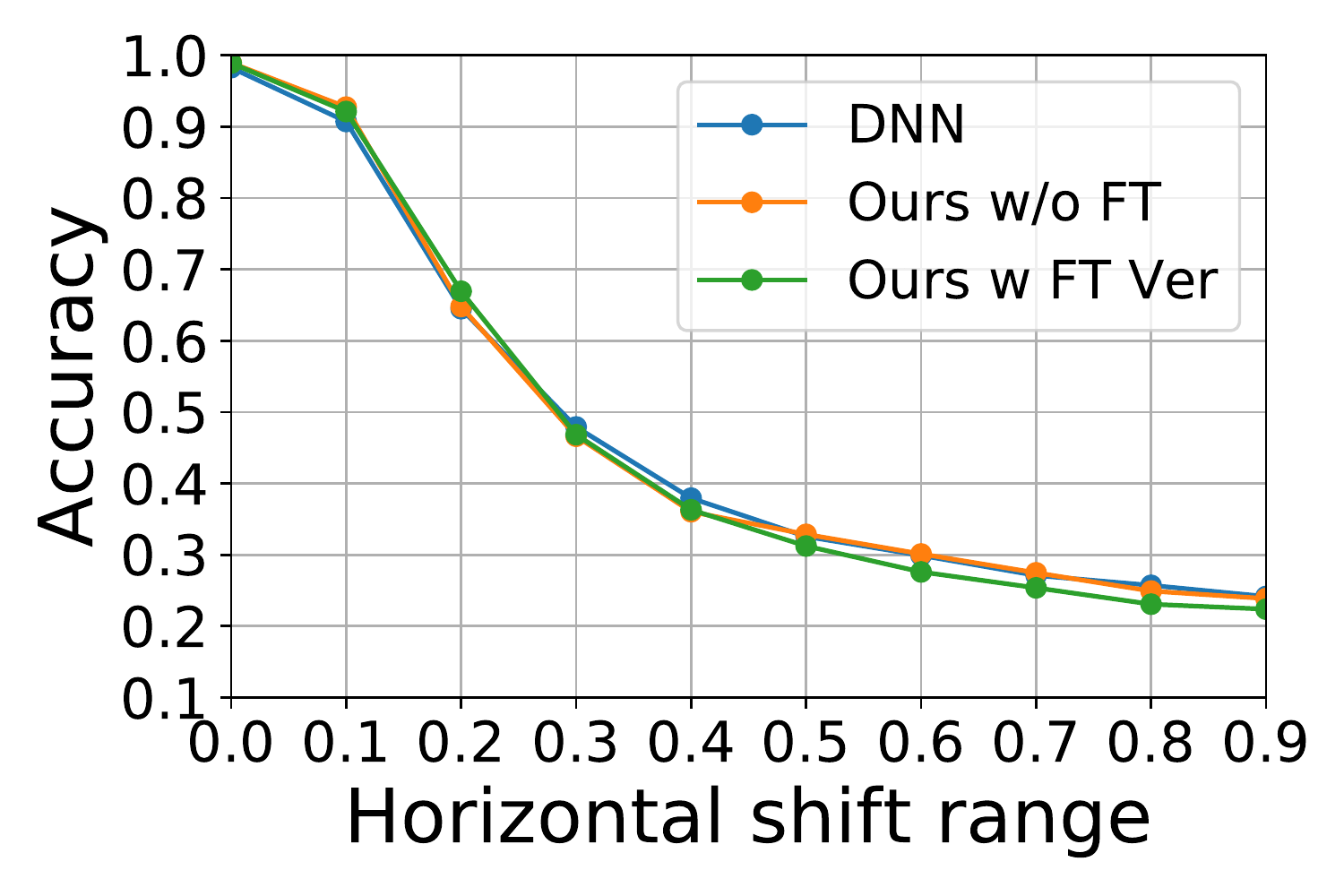}
\label{fig:VTHT}
}
\subfigure[Finetune HT test HT]{
\includegraphics[width=0.3 \textwidth]{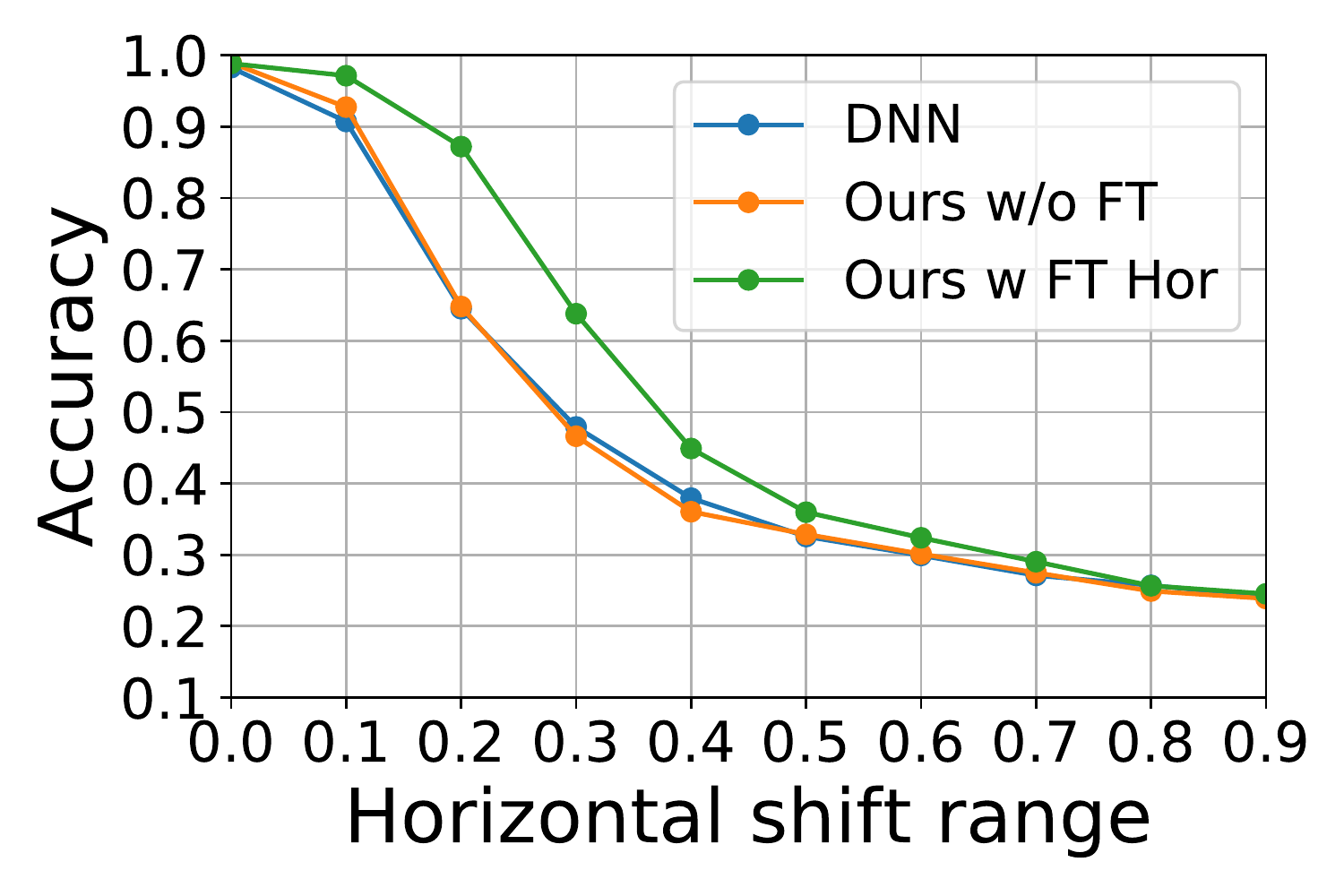}
\label{fig:HTHT}
}
\subfigure[Finetune Both test HT]{
\includegraphics[width=0.3 \textwidth]{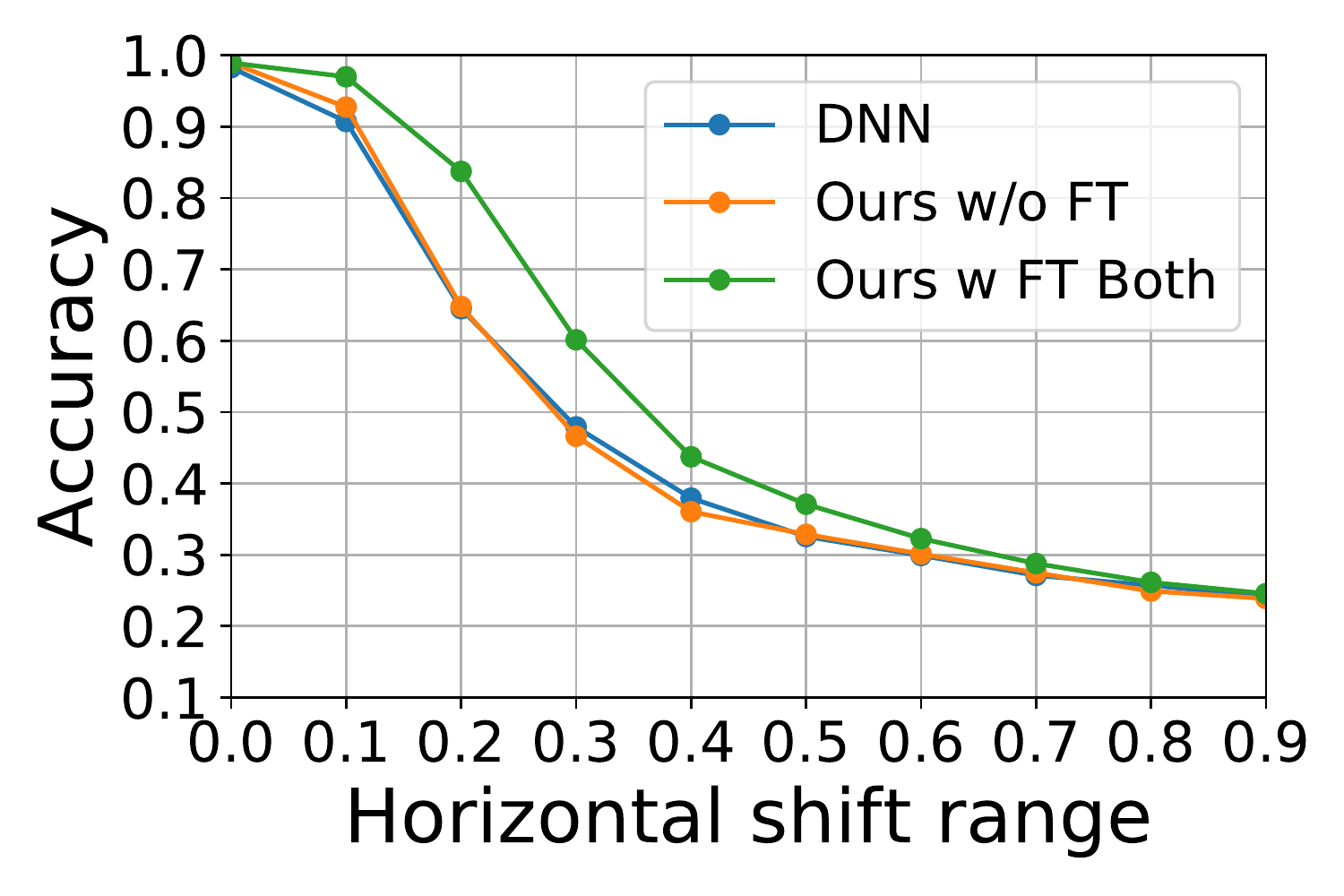}
\label{fig:BothHT}
}
\caption{Model robustness results on horizontal shifts}
\label{fig:train_clean_mnist_hor}
\end{figure*}

We present the test result regarding the horizontal shifts in Figure \ref{fig:train_clean_mnist_hor} which is under the same setting as Figure \ref{fig:train_clean_mnist} . The results support the same conclusion: fine-tuning on  data with different manipulation does not decrease the generalization ability with our model, which is shown in Figure \ref{fig:VTHT} where the green line (fine-tuning with vertical shifts) and the orange line (without fine-tuning) overlaps; Fine-tuning in testing time with the test data without label significantly improves the performance which is shown in  Figure \ref{fig:HTHT} and \ref{fig:BothHT} comparing to the orange line and the green line. Note that we have no intention to claim superior robustness of generative models to unseen manipulations without fine-tuning. Instead, we would like to show that they are able to obtain competitive performance when compared with discriminative models (gren vs blue lines). Our observation is consistent with previous work that when robustness is concerned, generative models are at least as competitive as discriminative ones  \cite{li2018generative}.  

\begin{figure}[]
\centering
\begin{minipage}[]{0.45\textwidth}
    \includegraphics[width= 0.9\textwidth]{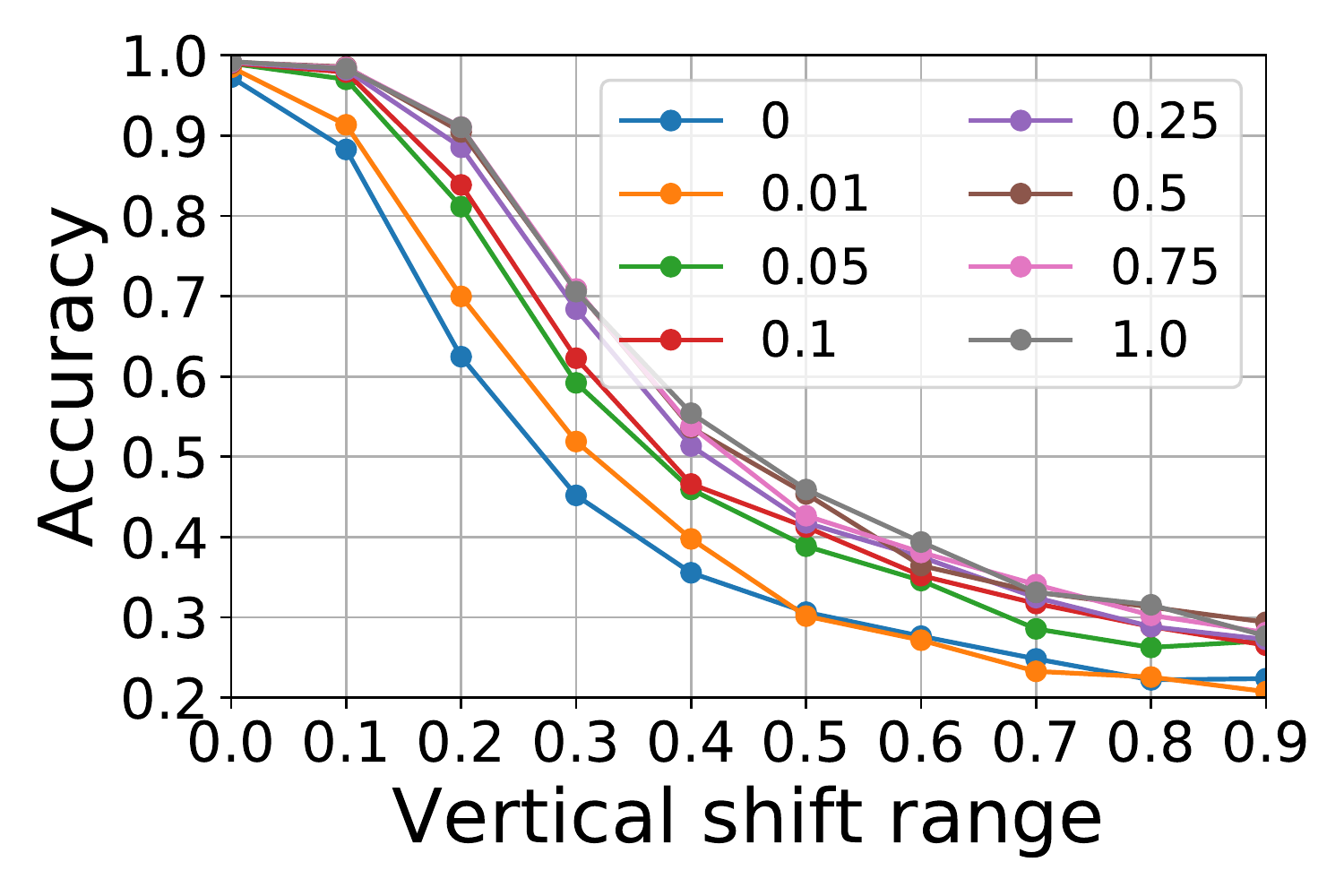}
    \caption{Performance regarding different percentage of test data used for fine-tuning manipulation of horizontal shift without using $do(m) = 0 $ for the cleaning training data during fine-tuning. }
    \label{fig:FT_percentage_notdoM_HT}
\end{minipage}
\hfill
\begin{minipage}[]{0.45\textwidth}
    \centering
    \includegraphics[width= 0.9\textwidth]{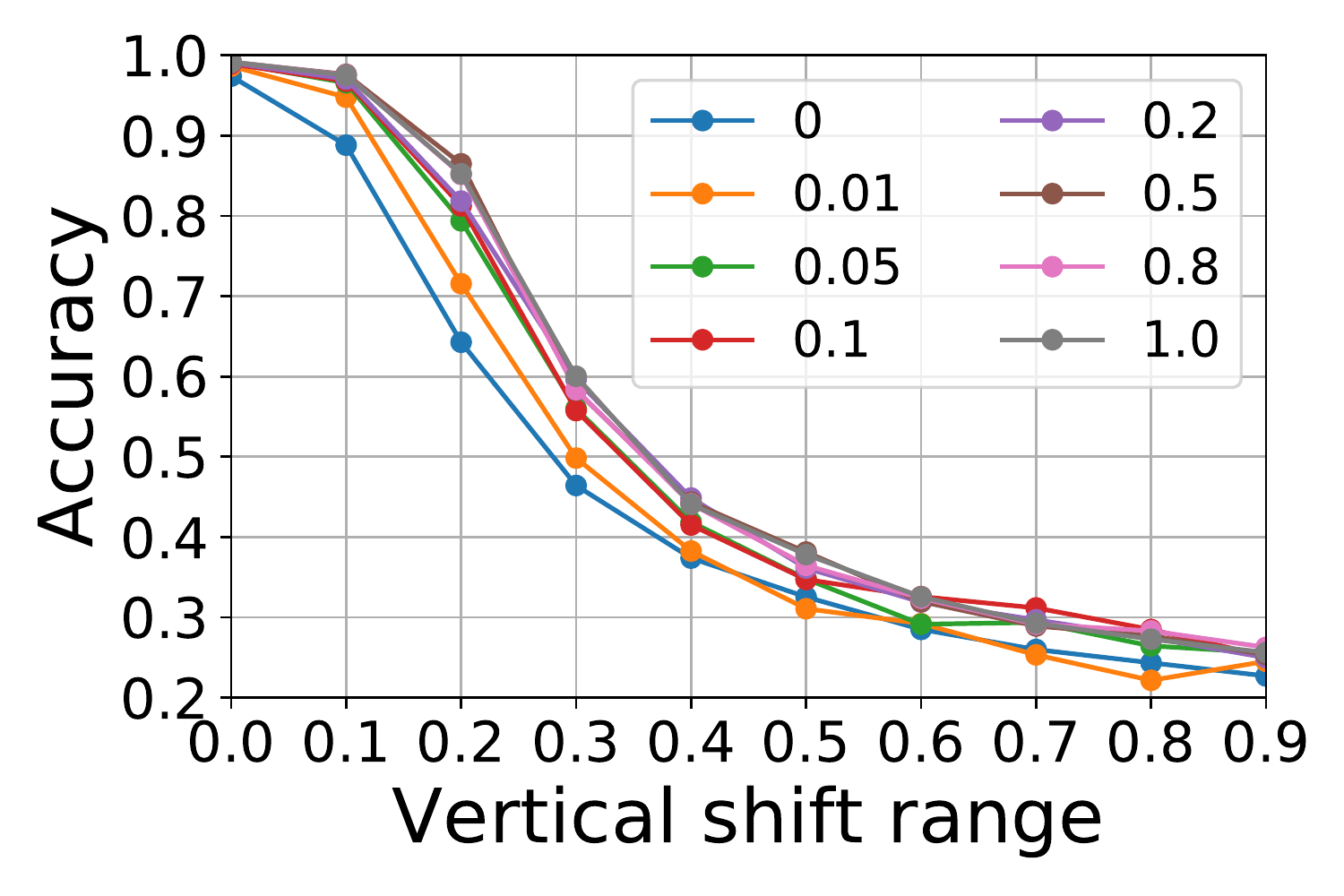}
    \caption{Performance regarding different percentage of test data used for fine-tuning manipulation  of vertical shift  using $do(m) = 0 $ for the cleaning training data during fine-tuning.}
    \label{fig:FT_percentage_notdoM}
\end{minipage}
\end{figure}

In addition to Figure \ref{fig:FT_percentage}, We also show the result testing with Vertical shift show in Figure \ref{fig:FT_percentage_notdoM_HT}, where a smaller $N_M^p$ network ([dimM, 500, 500]) is used. The conclusion is the same was using the vertical shift. We need very few data for fine-tune. More than $1\%$ data is sufficient.

Similar as Figure \ref{fig:FT_percentage}, we show the result using different percentage of data for fine-tuning in this experiment setting in  \ref{fig:FT_percentage_notdoM}.

\vspace{-7pt}
\subsection{Adversarial attack test on natural image classification}
\label{app:CIFAR}

%\begin{minipage}[]{0.49\textwidth}
%\begin{wrapfigure}{r}{0.55\linewidth}
\begin{figure}
\centering
\subfigure[FGSM]{
    \includegraphics[width=0.4\linewidth, height=0.3\linewidth]{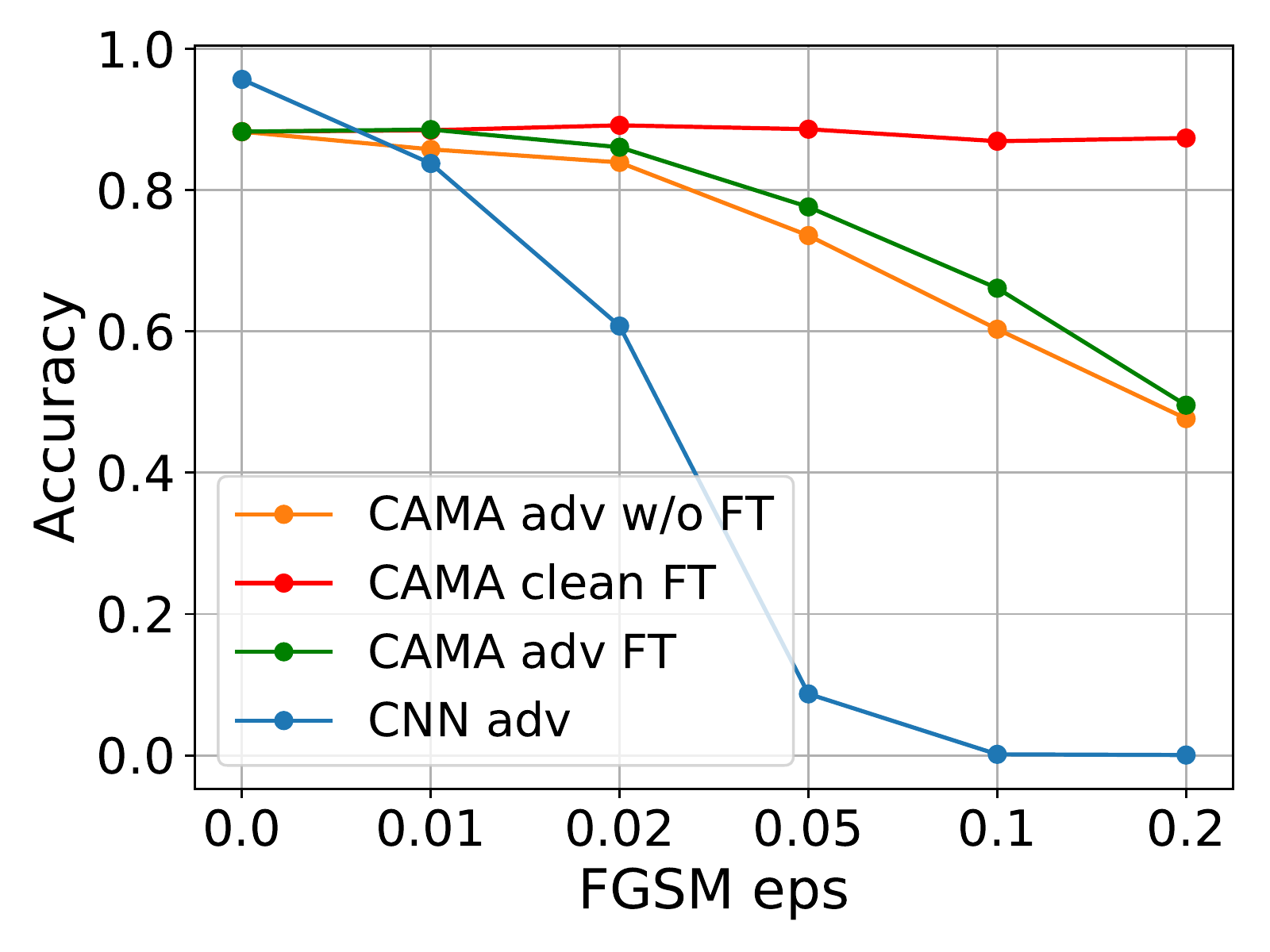}
    }
\subfigure[PGD]{
    \includegraphics[width=0.4\linewidth, height=0.3\linewidth]{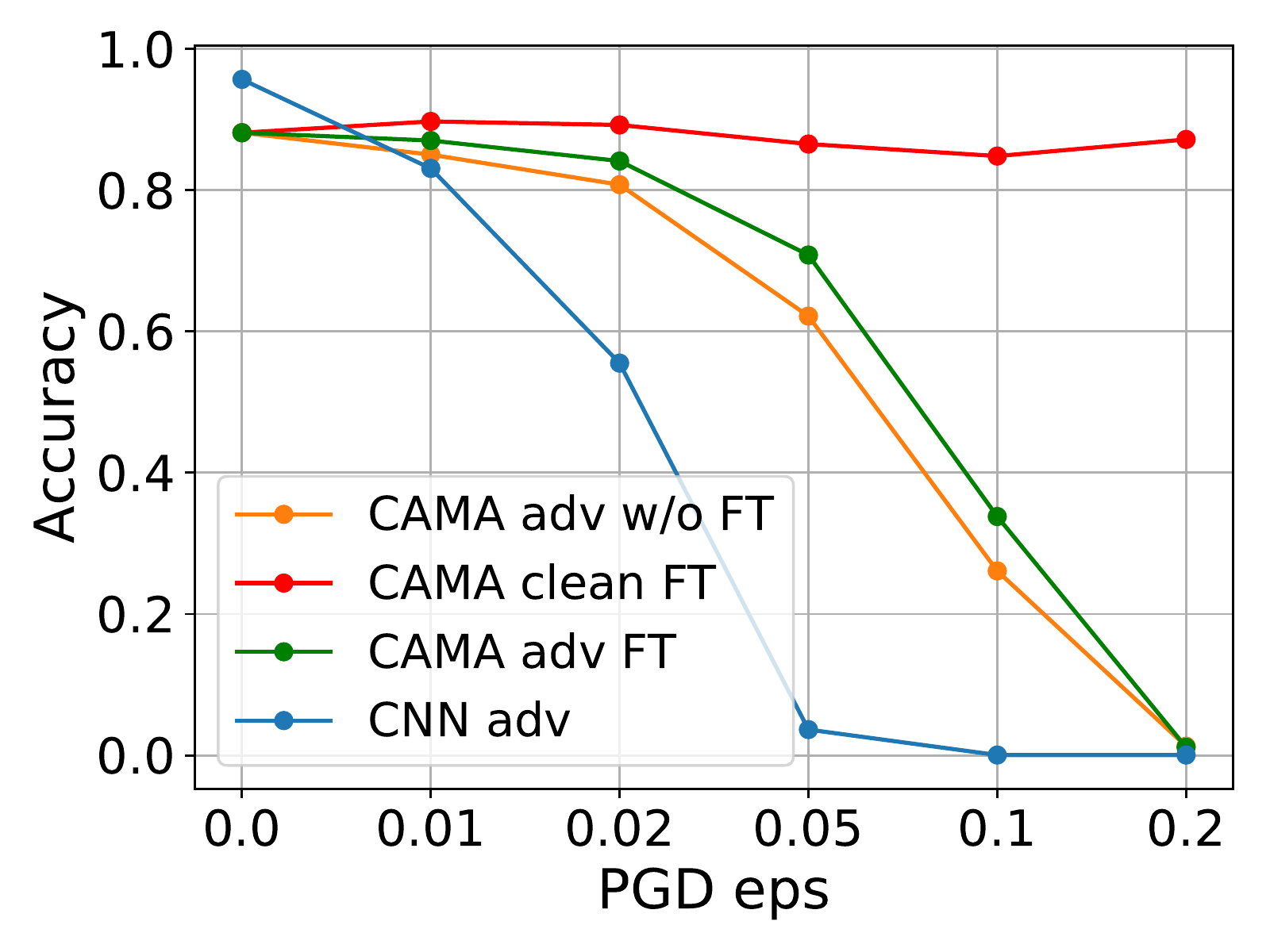}
    }
\caption{Adversarial robustness results on CIFAR-binary.}
\label{fig:CIFAR2_adversarial}
%\end{minipage}
\end{figure}
We evaluate the adversarial robustness of deep CAMA when trained on natural images. In this case we follow \citet{li2018generative} and consider \emph{CIFAR-binary}, a binary classification dataset containing “airplane” and “frog” images from CIFAR-10. We choose to work with CIFAR-binary because VAE-based fully generative classifiers are less satisfactory for classifying clean CIFAR-10 images ($< 50\%$ clean test accuracy). The deep CAMA model trained with data augmentation (adding Gaussian noise with standard deviation 0.1, see objective (\ref{eq:training_data_aug})) achieves $88.85\%$ clean test accuracy on CIFAR-binary, which is on par with the results reported in \citet{li2018generative}. For reference, a discriminative CNN with $2\times$ more channels achieves $95.60\%$ clean test accuracy. Similar to previous sections we apply FGSM and PGD attacks with different $\epsilon$ values to both deep CAMA and the discriminative CNN, and evaluate classification accuracies on the adversarial examples before and after finetuning. 

Results are reported in Figure \ref{fig:CIFAR2_adversarial}. For both FGSM and PGD tests, we see that deep CAMA, before finetuning, is significantly more robust to adversarial attacks when compared with a discriminative CNN model. Regarding finetuning, although PGD with large distortion ($\epsilon=0.2$) also fools the finetuning mechanism, in other cases finetuning still provides modest improvements ($5\%$ to $8\%$ when compared with the vanilla deep CAMA model) without deteriorating test accuracy on clean data. Combined with adversarial robustness results on MNIST, we conjecture that with a better generative model on natural images the robustness of deep CAMA can be further improved.

\subsection{Additional Baselines}
\label{sec:morebaseline}
%\begin{wrapfigure}[8]{l}{0.44\textwidth}
\begin{figure}
\centering
    %\vspace{-16pt}
    \hspace{-2pt}
    \includegraphics[width=7cm]{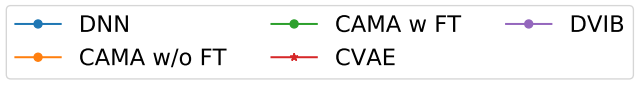}\\
    \vspace{-3pt}
    \hspace{-5pt}
    \includegraphics[height=4cm]{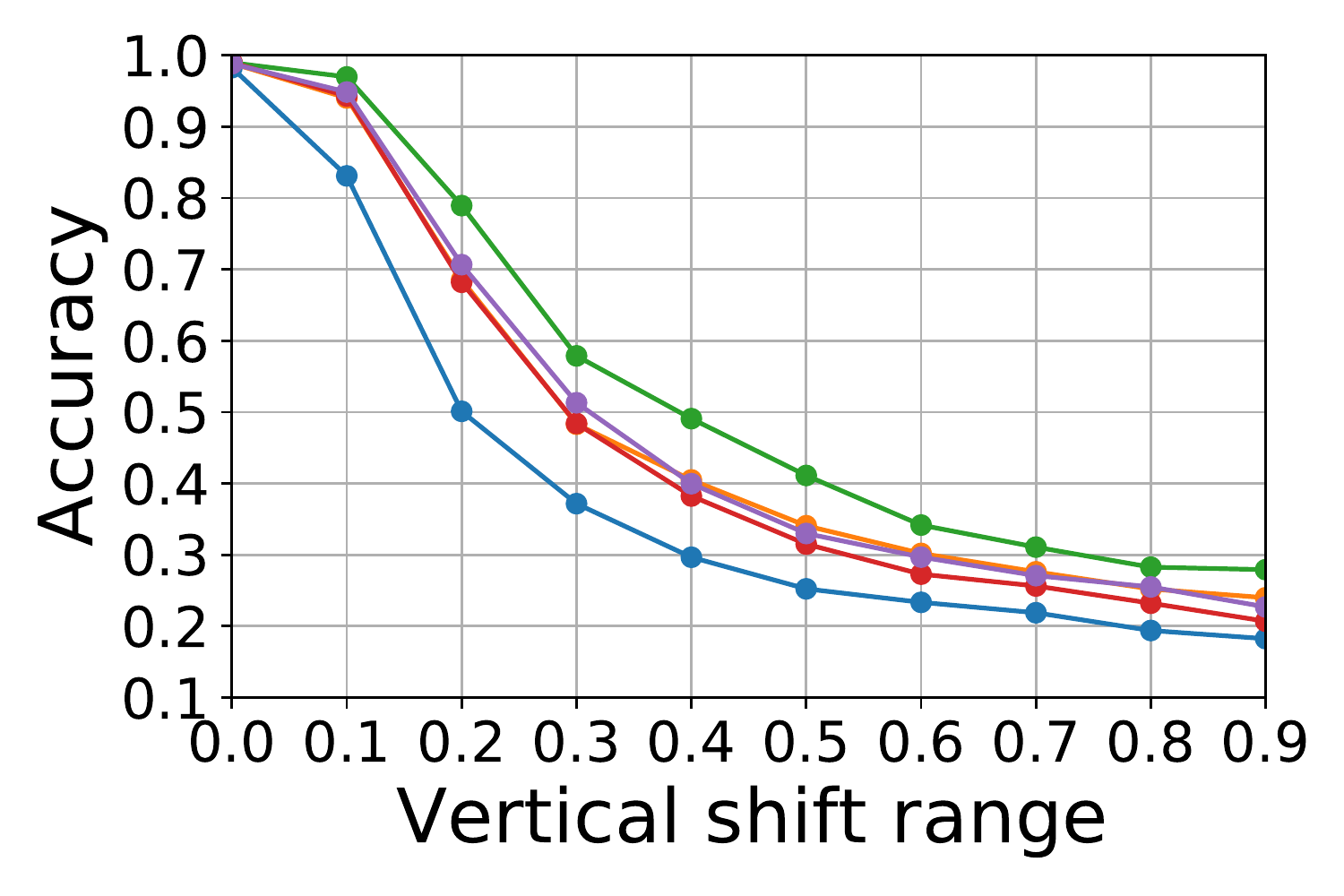}
    \hspace{-10pt}
    \includegraphics[height=4cm]{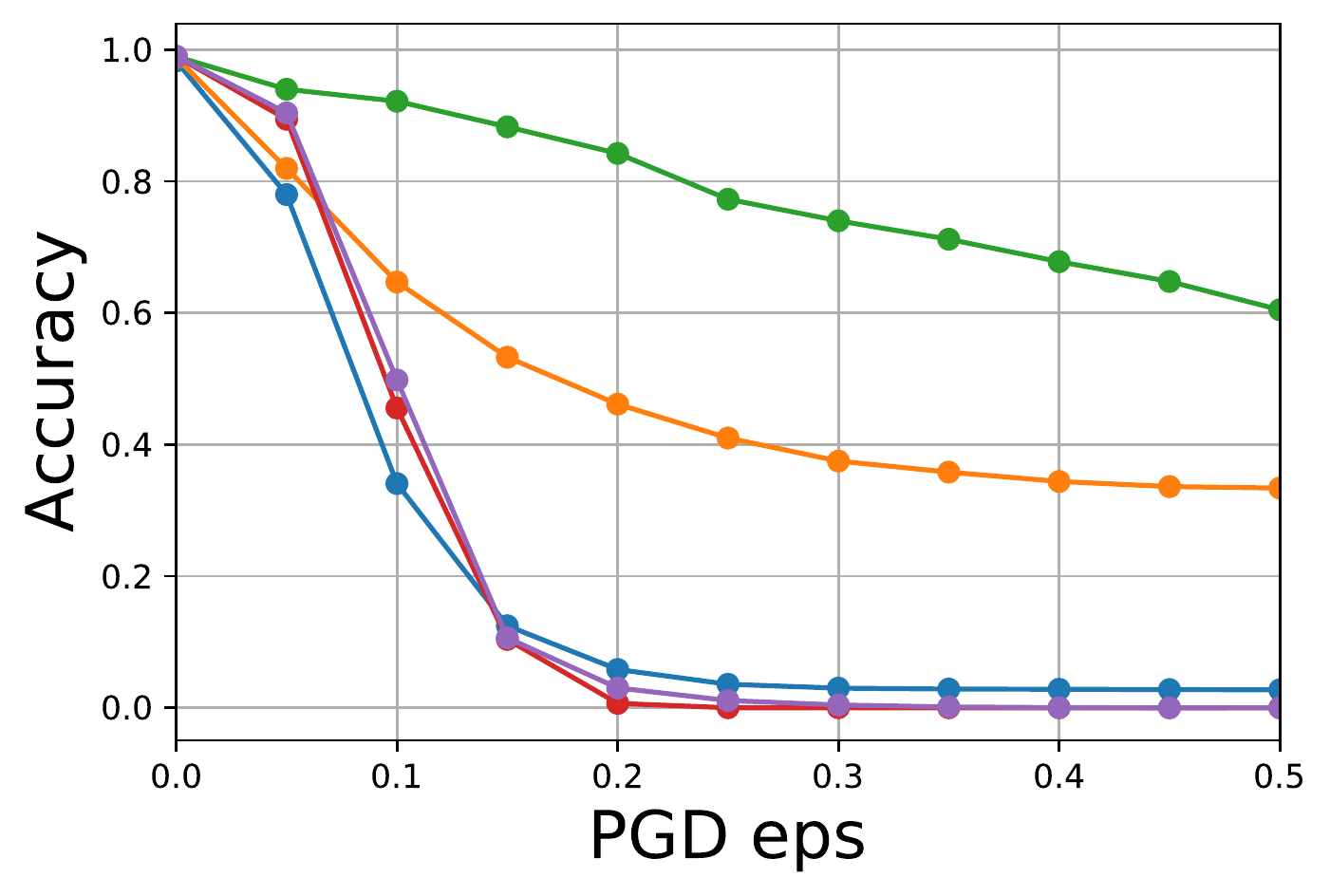}
    \hspace{-5pt}
    \caption{Comparisons to classifiers based on other deep latent variable models, including CVAE \citep{sohn2015learning} and DVIB \citep{alemi2016deep}. This again shows the benefit of the generative model structure in CAMA. }
    \label{fig:baseline}
\end{figure}
we present in Figure \ref{fig:baseline} the robustness of CVAE \citep{sohn2015learning} \& deep variational information bottleneck (DVIB) \citep{alemi2016deep} to vertical shifts and PGD attacks. Both baselines use \emph{discriminative} latent variable models for classification, and they perform either close to CAMA in vertical shift tests, or much worse than CAMA in PGD attack tests. This clearly shows the advantage of CAMA as a generative classifier, especially with fine-tuning.
Similar results have been reported in \citet{li2018generative}, and our work provides extra advantages due to the use of a causally consistent model and the fine-tuning method motivated by causal reasoning.

\subsection{Additional results with measurement data}
\label{sec:app_measurement}
\begin{figure}
     \subfigure[Manipulate co-parents $C$. ]{   \includegraphics[width=0.48\linewidth,height = 0.3\linewidth]{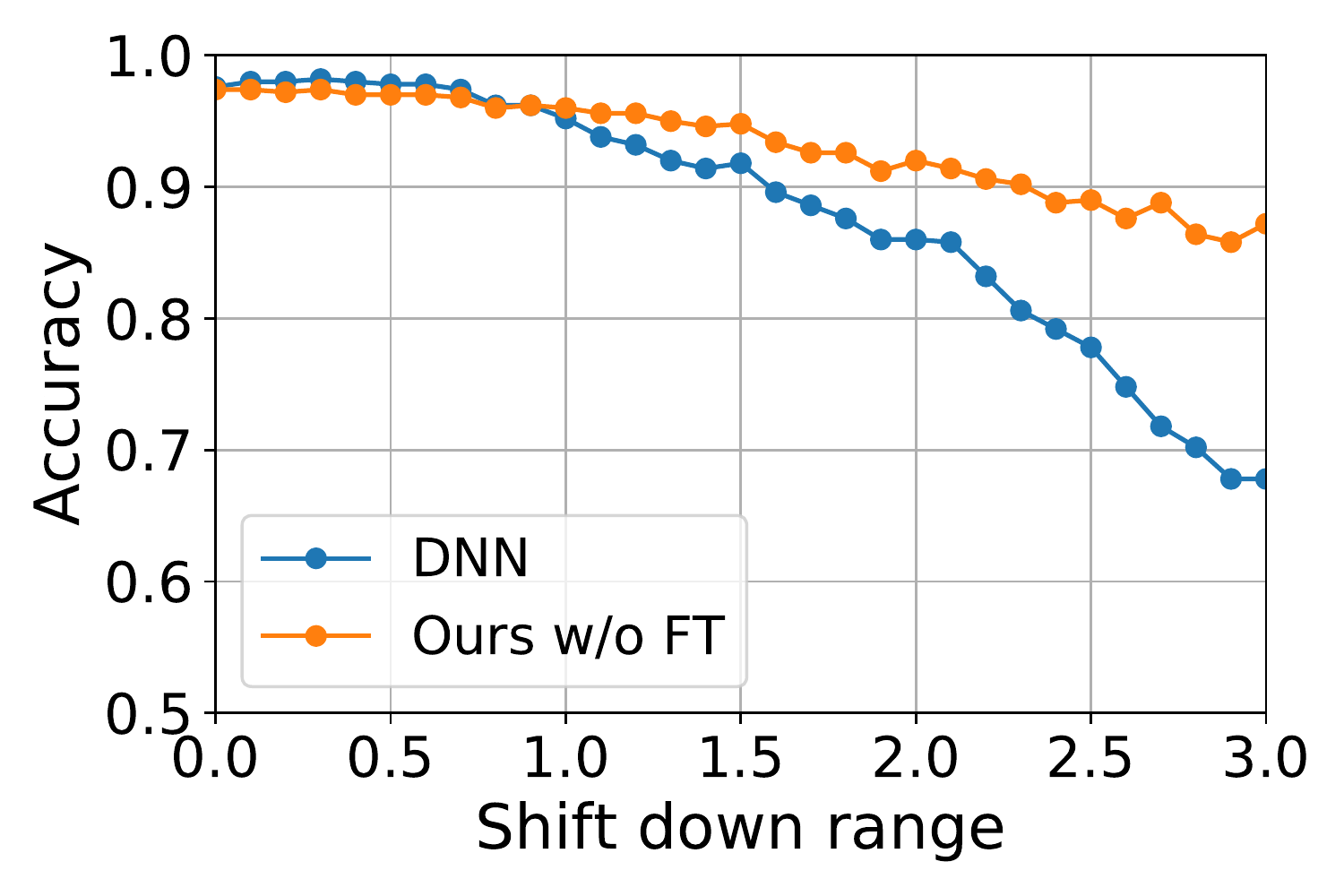}
    }
     \subfigure[Manipulate children $X$.]{   \includegraphics[width=0.48\linewidth,height = 0.3\linewidth]{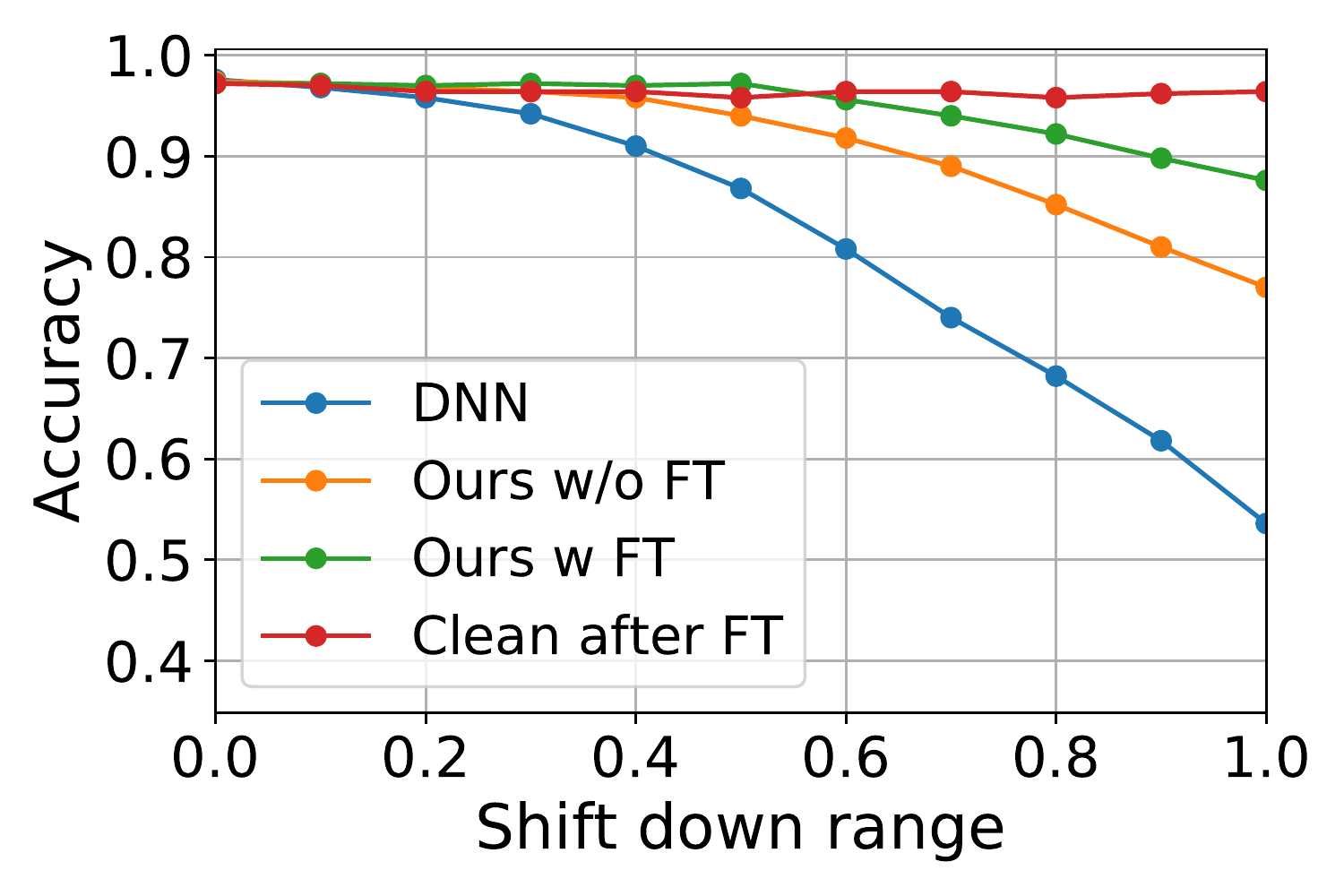}
    }
%     \vspace{-5pt}
     \caption{Shift Downscale}
%     %\vspace{-11pt}
     \label{fig:measurement_shiftdown}
\end{figure}

\paragraph{Shifting tests}
Figure \ref{fig:measurement_shiftdown} demonstrates the results on shifting the selected variables down-scale. The result is consistent with the shifting up-scale scenario in the main text. Deep CAMA is significantly more robust than baseline methods in both cases of manipulating co-parents and manipulating children. 

\begin{figure}
\vspace{-10pt}
    \centering
    \includegraphics[width= 0.9\linewidth]{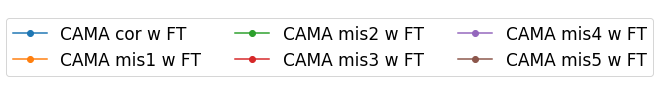}
    \includegraphics[width=0.45\linewidth]{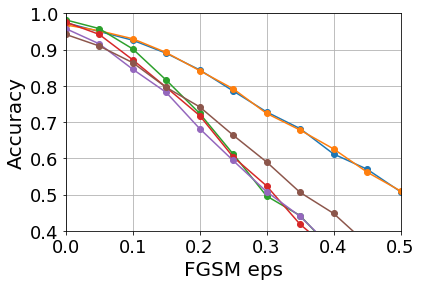}
    \includegraphics[width=0.45\linewidth]{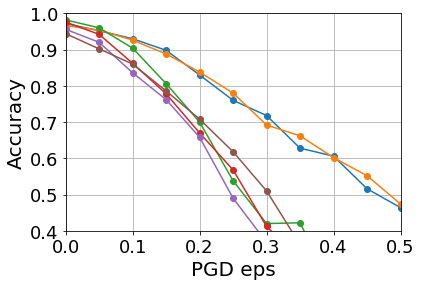}
       \vspace{-5pt}
    \caption{Test accuracy on measurement when causal relationship assumptions are violated. Swapping experiment. }
    \label{fig:SWAP_ChildCoP}
\end{figure}

% \begin{wrapfigure}[8]{r}{0.55\linewidth}
%     \centering
%     \vspace{-40pt}
%     \subfigure[Shift Upscale]{   \includegraphics[width=0.48\linewidth,height = 2.5cm]{Measurement/Measurement_ShiftUp_bigM_checkpoint9_doMTrue_FTSharedFalse.pdf}
%     }
%     \hspace{-15pt}
%      \subfigure[Shift Downscale]{   \includegraphics[width=0.48\linewidth,height = 2.5cm]{Measurement/Measurement_ShiftDown_bigM_checkpoint9_doMTrue_FTSharedFalse.pdf}
%     }
%     \vspace{-5pt}
%     \caption{Manipulate children $X$.}
%      \label{fig:manipulate_chilren}
% \end{wrapfigure}
\paragraph{Additional Experiments regarding assumption violation}
\label{app:app_violate}
We consider a more challenging scenario of a even more serious violation of the causal consistency assumption. In this case deep CAMA's graphical model is mis-specified by swapping the variables in the children/co-parent positions of the ground truth causal graph. Thus, in this case, the "mis1" experimental setting means two nodes are in the wrong positions (compared to the main text experiments where only one node is specified in the wrong position). Figure \ref{fig:SWAP_ChildCoP} shows the result of this co-parents/children swapping experiment. When only one pair of the variables is swapped (mis1, orange), the performance is still competitive with the causal consistent version of CAMA (cor). However, as we increase the number of swapped nodes, the robustness performance becomes significantly worse. This is again shows the importance of causal consistency for model design; still deep CAMA remains to be reasonably robust in the case of minor mis-specifications.

\section{Experimental settings}
\label{app:exp_settings}
\paragraph{Network architecture}
\begin{itemize}
    \item MNIST experiments:
    \begin{itemize}
        \item Discriminative DNN: The discriminate model used in the paper contains 4 densely connected hidden layer of $[512, 256, 126, 512]$ width for each layer. ReLU activations and dropout are used with dropout rate $[ 0.25, 0.25, 0.25, 0.5]$ for each layer.
        \item Deep CAMA's $p$ networks: we use $\text{dim}(Y) = 10, \text{dim}(Z) = 64$ and $\text{dim}(M) = 32$. \\
        $\text{NN}_Y^p$: an MLP of layer sizes $[\text{dim}(Y), 500, 500]$ and ReLU activations. \\
        $\text{NN}_Z^p$: an MLP of layer sizes $[\text{dim}(Z), 500, 500]$ and ReLU activations. \\
        $\text{NN}_M^p$: an MLP of layer sizes $[\text{dim}(M), 500, 500, 500, 500]$ and ReLU activations. \\
        $\text{NN}_{\text{merge}}^p$: an projection layer which projects the feature outputs from the previous networks to a 3D tensor of shape $(4, 4, 64)$, followed by 3 deconvolutional layers with stride 2, SAME padding, filter size $(3, 3, 64, 64)$ except for the last layer $(3, 3, 64, 1)$. All the layers use ReLU activations except for the last layer, which uses sigmoid activation. 
        \item Deep CAMA's $q$ networks: \\
        $\text{NN}_M^q$: it starts from a convolutional neural network (CNN) with 3 blocks of $\{ \text{conv} 3 \times 3, \text{max-pool}  \}$ layers with output channel size 64, stride 1 and SAME padding, then performs a reshape-to-vector operation and transforms this vector with an MLP of layer sizes $[4 \times 4 \times 64, 500, \text{dim}(M) \times 2]$ to generate the mean and log-variance of $q(m | x)$. All the layers use ReLU activation except for the last layer, which uses linear activation.  \\
        $\text{NN}_Z^q$: first it uses a CNN with similar architecture as $\text{NN}^M_q$'s CNN (except that the filter size is 5) to process $x$. Then after the reshape-to-vector operation, the vector first gets transformed by an MLP of size $[4 \times 4 \times 64, 500]$, then it gets combined with $y$ and $m$ and passed through another MLP of size $[500 + \text{dim}(Y) + \text{dim}(M), 500, \text{dim}(Z) \times 2]$ to obtain the mean and log-variance of $q(z|x, y, m)$. All the layers use ReLU activation except for the last layer, which uses linear activation.
    \end{itemize}
    \item Measurement data experiments:
    \begin{itemize}
        \item Discriminative DNN: 
        The $A, C, X$ variables are concatenated to an input vector of total dimension $20$. Then the DNN contains 3 densely connected hidden layer of $[64, 16, 32]$ width for each layer, and output  $Y$. ReLU activations and dropout are used with dropout rate $[ 0.25, 0.25, 0.5]$ for each layer.
        \item Deep CAMA's $p$ networks: we use $\text{dim}(Y) = 5, \text{dim}(A) = 5, \text{dim}(C) = 5, \text{dim}(Z) = 64$ and $\text{dim}(M) = 32$. \\
        $p(y|a)$: an MLP of layer sizes $[\text{dim}(A), 500, 500, \text{dim}(Y)]$, ReLU activations except for the last layer (softmax). \\
        $p(x|y, c, z, m)$ contains 5 networks: 4 networks $\{ \text{NN}_Y^p, \text{NN}_C^p, \text{NN}_Z^p, \text{NN}_M^p  \}$ to process each of the parents of $X$, followed by a merging network. \\
        $\text{NN}_Y^p$: an MLP of layer sizes $[\text{dim}(Y), 500, 500]$ and ReLU activations. \\
        $\text{NN}_C^p$: an MLP of layer sizes $[\text{dim}(C), 500, 500]$ and ReLU activations. \\
        $\text{NN}_Z^p$ an MLP of layer sizes $[\text{dim}(Z), 500, 500]$ and ReLU activations. \\
        $\text{NN}_M^p$: an MLP of layer sizes $[\text{dim}(M), 500, 500, 500, 500]$ and ReLU activations. \\
        $\text{NN}_{\text{merge}}^p$: it first start from a concatenation of the feature outputs from the above 4 networks, then transforms the concatenated vector with an MLP of layer sizes $[500 \times 4, 500, \text{dim}(X)]$ to output the mean of $x$. All the layers use ReLU activations except for the last layer, which uses linear activation.
        \item Deep CAMA's $q$ networks: \\
        $q(m|x)$: it uses an MLP of layer sizes $[\text{dim}(X), 500, 500, \text{dim}(M) \times 2]$ to obtain the mean and log-variance. All the layers use ReLU activations except for the last layer, which uses linear activation. \\
        $q(z | x, y, m, a, c)$: it first concatenates $x, y, m, a, c$ into a vecto, then uses an MLP of layer sizes $[\text{dim}(X) + \text{dim}(Y) + \text{dim}(M) + \text{dim}(A) + \text{dim}(C), 500, 500, \text{dim}(Z) \times 2]$ to transform this vector into the mean and log-variance of $q(z|x, y, m, a, c)$. All the layers use ReLU activations except for the last layer, which uses linear activation.
    \end{itemize}
    \item CIFAR-binary experiments:
    \begin{itemize}
        \item Discriminative CNN: The discriminate model used in the paper is a CNN with 3 convolutional layers of filter width 3 and channel sizes [128, 128, 128], followed by a flattening operation and a 2-hidden layer MLP of size [$4 \times 4 \times 128, 1000, 1000, 10$]. It uses ReLU activations and max pooling for the convolutional layers.
        \item Deep CAMA's $p$ networks: we use $\text{dim}(Y) = 10, \text{dim}(Z) = 128$ and $\text{dim}(M) = 64$. \\
        $\text{NN}_Y^p$: an MLP of layer sizes $[\text{dim}(Y), 1000, 1000]$ and ReLU activations. \\
        $\text{NN}_Z^p$: an MLP of layer sizes $[\text{dim}(Z), 1000, 1000]$ and ReLU activations. \\
        $\text{NN}_M^p$: an MLP of layer sizes $[\text{dim}(M), 1000, 1000, 1000]$ and ReLU activations. \\
        $\text{NN}_{\text{merge}}^p$: an projection layer which projects the feature outputs from the previous networks to a 3D tensor of shape $(4, 4, 64)$, followed by 4 deconvolutional layers with stride 2, SAME padding, filter size $(3, 3, 64, 64)$ except for the last layer $(3, 3, 64, 3)$. All the layers use ReLU activations except for the last layer, which uses sigmoid activation. 
        \item Deep CAMA's $q$ networks: \\
        $\text{NN}_M^q$: it starts from a convolutional neural network (CNN) with 3 blocks of $\{ \text{conv} 3 \times 3, \text{max-pool}  \}$ layers with output channel size 64, stride 1 and SAME padding, then performs a reshape-to-vector operation and transforms this vector with an MLP of layer sizes $[4 \times 4 \times 64, 1000, 1000, \text{dim}(M) \times 2]$ to generate the mean and log-variance of $q(m | x)$. All the layers use ReLU activation except for the last layer, which uses linear activation.  \\
        $\text{NN}_Z^q$: first it re-uses $\text{NN}_M^q$ CNN network for feature extraction on $x$. Then after the reshape-to-vector operation, the vector gets combined with $y$ and $m$ and passed through another MLP of size $[4 \times 4 \times 64 + \text{dim}(Y) + \text{dim}(M), 1000, 1000, \text{dim}(Z) \times 2]$ to obtain the mean and log-variance of $q(z|x, y, m)$. All the layers use ReLU activation except for the last layer, which uses linear activation.
    \end{itemize}
\end{itemize}

\paragraph{Measurement data generation}
We set the target $Y$ to be categorical, its children, co-parents and parents are continuous variables.
The set $5$ classes for $Y$, and $Y$ has $10$ children variables and $5$ co-parents variables, also one $5$ dimensional parents. 

Parents ($A$) and co-parents ($C$) are generated by sampling from a normal distribution. 
We generate $Y$ using structured equation $Y= f_y(A) + \sigma_Y$. We use $f_y= \text{argmax} ~ g(A)$ and $g()$ is a quadratic function $0.2*A^2 -0.8A$. $\sigma_Y$ is the Gaussain noise. 

To generate the children $X= f(Y, C) + \sigma_x$, we also used quadratic function $f$ and the parameters were sampled from a Gaussian distribution. As in the experiment, we were using fixed scale shift, we also added a normalize the children before adding the Gaussian random noise $\sigma_x$. So that all observations are in similar scale.

\paragraph{Other}
For MNIST experiments, we uses $5\%$ of the training data as the validation set. 
We used the training results with the highest validation accuracy for testing. 
 If not otherwise specified, $50\%$ of noisy test data are used for fine-tuning in the shift experiments and all data are used for fine-tuning in the attack experiments.

For the experiments with measurement data. We generated $1000$ data in total. We split, $500$ data for testing, $450$ for training and $50$ for validation. We used the training results with the highest validation accuracy for testing for both deep CAMA and for DNN.